\documentclass[sigconf,nonacm]{acmart}

\AtBeginDocument{%
  \providecommand\BibTeX{{%
    \normalfont B\kern-0.5em{\scshape i\kern-0.25em b}\kern-0.8em\TeX}}}

\setcopyright{none}
\copyrightyear{ }
\acmYear{ }
\acmDOI{}
\newsavebox\ltmcbox

\usepackage{lineno,hyperref}
\usepackage{longtable}
\usepackage{graphicx}
\usepackage{multirow}
\usepackage{enumitem}
\usepackage{amsfonts}

\usepackage[ruled,vlined,linesnumbered]{algorithm2e}
\usepackage{color}

\usepackage{multirow}
\usepackage{enumitem}
\usepackage{cases}
\usepackage{subeqnarray}
\usepackage{subfigure}
\usepackage{bbding}

\usepackage{wrapfig}

\acmBooktitle{}
\acmPrice{}
\acmISBN{}

\begin{document}

\title{Multi Loss-based Feature Fusion and Top Two Voting Ensemble Decision Strategy for Facial Expression Recognition in the Wild}

\author{Guangyao Zhou}
\email{guangyao\_zhou@std.uestc.edu.cn}
\affiliation{%
  \institution{School of Computer and Artificial Intelligence, Southwest Jiaotong University, China.}
  \city{Cheng Du}
  \country{China}
}
\author{Yuanlun Xie}
\email{ylxie@std.uestc.edu.cn}
\affiliation{%
  \institution{Chengdu University, China}
  \city{Cheng Du}
  \country{China}
}

\author{Yiqin Fu}
\affiliation{%
  \institution{China National Offshore Oil Corporation, China}
  \city{Beijing}
  \country{China}
}
\author{Zhaokun Wang}
\affiliation{%
  \institution{School of Information and Software Engineering, University of Electronic Science and Technology of China}
  \city{Cheng Du}
  \country{China}
}

\renewcommand{\shortauthors}{ }

\begin{abstract}
Facial expression recognition (FER) in the wild is a challenging task affected by the image quality and has attracted broad interest in computer vision. There is no research using feature fusion and ensemble strategy for FER simultaneously. Different from previous studies, this paper applies both internal feature fusion for a single model and feature fusion among multiple networks, as well as the ensemble strategy. This paper proposes one novel single model named R18+FAML, as well as one ensemble model named R18+FAML-FGA-T2V to improve the performance of the FER in the wild. Based on the structure of ResNet18 (R18), R18+FAML combines internal Feature fusion and three Attention blocks using Multiple Loss functions (FAML) to improve the diversity of the feature extraction. To improve the performance of R18+FAML, we propose a Feature fusion among networks based on the Genetic Algorithm (FGA), which can fuse the convolution kernels for feature extraction of multiple networks. On the basis of R18+FAML and FGA, we propose one ensemble strategy, i.e., the Top Two Voting (T2V) to support the classification of FER, which can consider more classification information comprehensively. Combining the above strategies, R18+FAML-FGA-T2V can focus on the main expression-aware areas. Extensive experiments demonstrate that our single model R18+FAML and the ensemble model R18+FAML-FGA-T2V achieve the accuracies of $\left( 90.32, 62.17, 65.83 \right)\%$ and $\left( 91.59, 63.27, 66.63 \right)\%$ on three challenging unbalanced FER datasets RAF-DB, AffectNet-8 and AffectNet-7 respectively, both outperforming the state-of-the-art results.
\end{abstract}


\maketitle

\begin{sloppypar}

\section{Introduction}\label{sec1}
Facial expression recognition (FER) has attracted increasing concerns in research fields of medical treatment, psychoanalysis and dialogue robot \cite{f2,f27}. As an approach to assist computers to comprehend human emotions and improve human-computer interaction, FER is regarded as an indispensable ability of computer vision \cite{f3}.

According to the environment of facial images, the existing datasets of FER can be divided into datasets in the laboratory and that in the wild.
Existing research on FER using deep learning technology mainly focuses on expression classification and has achieved significant performance.
Currently, some FER models including Auto-FERNet \cite{f2}, deep-learned Tandem Facial Expression (TFE) \cite{f4}, OAENet \cite{f5} and FER-VT \cite{f6} have almost achieved a top accuracy more than 98\% in some laboratory datasets such as CK+, JAFFE.  However, FER is still a complex task for wild datasets collected from the real world due to four factors:
1) illumination, different optical parameters affect the overall value of the three-channel data of the image;
2) occlusion, such as that part of the face may be occluded by hands;
3) large inter-class similarities, the differences between expressions from different classes are small, making it difficult to distinguish;
4) unbalanced data amount between classes, the amount of training data in different classes varies greatly.

These factors are eventually challenging expression recognition for some wild datasets such as RAF-DB \cite{f7}, ExpW \cite{f10}, EmotioNet \cite{f9} and AffectNet \cite{f8}. Additionally, the division of FER datasets depends on the subjective judgement of experts resulting in inconsistent boundaries of some expressions \cite{f1}.

In image recognition, VGGNet \cite{f11}, ResNet \cite{f12} and DenseNet \cite{f13} are frequently used deep models to extract the feature of image and have shown remarkable performance in the image recognition. However, there is no specialized model for feature extraction of FER, especially for wild unbalanced datasets \cite{f2}. Facial expression is the presentation of human emotion varying with the ages, races and cultural backgrounds, which also increase the difficulty of FER in reality. Thus, feature extraction of expression requires a more efficient model to enhance the accuracy of FER, especially for wild datasets.

Most studies of FER in the wild datasets usually learn facial expression features from the specific information of the facial expression image itself.
For example, DisEmoNet \cite{f14} extracted affect-salient features by disentangling emotion-related factors from other factors such as head poses, illuminations and occlusions; multiscale graph convolutional network (GCN) \cite{f15} extracted landmark graphs as a type of facial expression features; as well as ADC-Net \cite{f16} learned more discriminative expression features from crucial local facial regions by partitioning the face image into several regions.
Additionally, with the utilization of the attention mechanism, TransFER model \cite{f17}, DAN \cite{f48} and MA-Net \cite{f25} obtained higher accuracies on several wild FER datasets than that of previous studies, which demonstrates the possibility to further improve the accuracy. Concurrently, it also illustrates the worth of exploration of features extraction and features optimization.
Although these methods have appreciable performance, they only leveraged gradient descent to update the weights of convolution kernels generally applying the single loss function to approximate the training labels, which easily results in the local optimal solution.

In the image recognition task, CNN (convolution neural network) is a usual method to extract image features that are actually characterized by the weights of convolution kernels \cite{f24,f23}. Therefore, for FER that belongs to image recognition, the research on optimizing the weights of convolution kernels is significant and conducive to ameliorating the effect of feature extraction.

Each strategy is the crystallization of human intelligence.
No strategy can replace another in all scenarios.
The incorporation of them is a better way to address complex tasks.
FER is a challenging task and improving the accuracy by 0.1\% is very difficult, e.g., when above 88\% in RAF-DB, etc.
Training DNNs is actually functional to search optimal parameter within the solution space limited by structure. This paper embarks on an exploration to more possibilities to further improve accuracy of FER, hoping to contribute beneficial insights to FER and DL.
In this paper, we focus on optimizing the weights of convolution kernel in FER models to improve the accuracy of FER. Based on this, we consider improving the accuracy of FER from three aspects: single model structure, interaction among multiple networks, ensemble decision making based on multiple networks.

For improvement of the single model structure, we propose a model named FAML combining internal Feature fusion and Attention mechanism based on the structure of ResNet18 and use {M}ultiple {L}oss functions to train the model. The utilization of multiple loss functions can jump out of the local convergence point of a single loss function.
For interaction among multiple networks, we propose the {F}eature fusion among networks based on the {G}enetic {A}lgorithm (FGA), which integrates the feature information of multiple networks to alleviate the overfitting so as to further improve the performance of the single model.
Finally, we propose one universal ensemble strategy, that {T}op {T}wo {V}oting (T2V), to support the decision-making of FER, which further enhances the accuracy of FER. The ensemble approach can reduce overfitting \cite{f18}. Some existing ensemble studies for FER include probability-based fusion \cite{f18}, multiple impression feedback recognition model (MIFR) with ensemble SVM \cite{f19}, smoothed deep neural network ensemble with bootstrap aggregation \cite{f20}, DCNN-SVM with bagging ensemble and DCNN-VC with voting technique \cite{f21}, as well as ensemble classifiers with employing Neural Network and SVM \cite{f22}.

Different from existing methods, our proposed ensemble strategy, i.e. T2V, considers the multilevel ranking in results of each network and directly participates in the decision of classification instead of participating in the training of each network. This idea comes from the phenomenon that in the experiments of FER, the secondary class of some networks' output on their false recognized sample data is probably the labeled category. Therefore, it can be regarded that the networks have been well trained, but they are still ``hesitant" in Top 1 and Top 2 for these sample data.
Generally, the existing studies are mainly based on the feature learning of a single network to improve performance. This paper not only proposes an efficient network but also considers fusing the features of multiple networks to further improve the performance of a single model and the decision results of overall expression recognition. Then, combining the above three aspects, we obtain a single model FAML and an ensemble model FAML-FGA-T2V that achieve state-of-the-art performance on three wild unbalanced FER datasets.

The main contributions of this paper can be summarized as follows.
\begin{enumerate}[label=(\arabic*), leftmargin=*, topsep = 0em, partopsep=0em, itemsep=0em, parsep = 0em]
\item Proposing an efficient model FAML combining Internal Feature fusion with Attention block using Multiple Loss functions of FER in the wild to improve the feature extraction of CNNs. Multi-loss strategy provides multi-search paths for training networks, so as to improve the diversity of feature extractions. In theory, the local optimum of multi-path search must satisfy the convergence conditions of all paths simultaneously, hence usually better than that of single-path.
\item Proposing fusing features among multiple networks based on the Genetic Algorithm (FGA) to evolve the feature extraction of every single network. The application of the genetic algorithm makes the descendant networks inherit the excellent convolution kernel weights of the parents, so as to keep the convolution kernels with better performance of feature extraction. 
\item Proposing the improved Top Two Voting (T2V) ensemble decision strategy to support the classification of FER. On the basis of multiple loss functions, improved T2V for FER can better integrate the interest areas of multiple networks with the same structure.    By combining FAML, FGA and T2V, we obtain a well-performed ensemble model named FAML-FGA-T2V, which provides a set of effective strategies to improve the accuracy of FER. 
\item Extensive Experiments demonstrate that our single model R18+FAML and the ensemble model R18+FAML-FGA-T2V achieve the accuracies of $(90.32,62.17,65.83)\%$  and $(91.59,63.27,66.63)\%$ on three challenging unbalanced FER datasets RAF-DB, AffectNet-8 and AffectNet-7 respectively, both outperforming the state-of-the-art results of single networks and ensemble FER models.
\end{enumerate}

The remainder of this paper is organized as follows.
We review the related work in Section \ref{sec2}. The network and methodologies including FAML, FGA and T2V ensemble strategy are proposed in Section  \ref{sec3}. The experiments procedure and results are presented in
Section \ref{sec4}. Finally, we conclude this paper in Section \ref{sec5}.

\section{Related Work}\label{sec2}
In this section, we review the related work about facial expression recognition, feature fusion and ensemble strategies briefly.
\subsection{Facial Expression Recognition}\label{sec2.1}
Facial expression recognition (FER) as a branch of image recognition has kept an attractive research direction in past decades. FER is generally based on image or video \cite{f25}. We mainly review the work of image-based FER that is consistent with the focus of this paper.

Generally, image-based FER has two types of datasets i.e. datasets in the lab and that in the wild \cite{f1}. Next, we will review the work on these two types of datasets respectively.

With higher quality expression images, the studies of the lab-dataset have recently reached a high level in these literature \cite{f4,f5,f6,f15,f27,f28,f29,f30,f31}. The methods in these literature can also reflect the researchers' exploration of FER.
Label distribution learning is to train a model of label distribution of learning samples to better represent the distribution of a sample.
Chen et al. \cite{f27} introduced intensity label distribution learning and proposed a joint FER and expression intensity estimation method by encoding the expression intensity in a multidimensional expression space to successfully analyze children's empathy ability, which was also validated on public datasets of FER.
The image reconstruction is to smoothly complete the occluded part through the lines of the unobstructed part to assist the image recognition.
Poux et al. \cite{f28} leveraged the generative algorithms to reconstruct occluded facial expressions by reconstructing directly the occluded optical flow and proposed a denoising auto-encoder trained to reconstruct corrupted optical flow recognition to solve FER in the presence of partial facial occlusions.
Landmark is to extract the key points of the face which can reduce the influence of individual characteristics.
Rao et al. \cite{f15} proposed a novel multiscale graph convolutional network (GCN) based on landmark graphs extracted from facial images and focus on the crucial parts that significantly
affect FER.
With a similar principle to the landmark graph, the local region is also a method to support FER by extracting partial information.
Karimi et al. \cite{f49} proposed a partially connected Multilayer Perceptron (PCM) neural network as an optimal new MLP to detect face emotions, which had faster convergence.
Jin et al. \cite{f28} proposed MiniExpNet, a lightweight network based on facial local regions by human-machine collaborative strategy, which provided an appreciable balance between accuracy, model size and speed. In \cite{f28}, MiniExpNet also applied the attention mechanism and distillation mechanism which are two frequency mechanisms in recent research to reduce overfitting and improve the accuracy of the model. Ma et al. \cite{f50} used multi-level knowledge distillation to detect low-resolution and improve FER performance.
For attention mechanism:
Sun et al. \cite{f33} proposed ASModel (Attention Shallow Model) by using the relative position of facial landmarks and the texture characteristics of the local area to describe the action units of the face;
Wang et al. \cite{f5} proposed OAENet (Oriented Attention Enable Network) aggregating attention mechanism and ensuring the sufficient utilization of both global and local features;
Shahid et al. \cite{k2} proposed SqueezExpNet (dual-stage CNN with attention mechanism) to enhance the accuracy of FER.
Huang et al. \cite{f6} proposed FER-VT with a grid-wise attention mechanism to capture the dependencies of different regions from a facial expression image, which achieved $100\%$ accuracy on CK+ dataset.
Other strategies for FER include Two-branch Disentangled Generative Adversarial
Network (TDGAN) \cite{f29}, deep reinforcement learning \cite{k4} adaptive weighting of handcrafted feature \cite{f30}, architecture search \cite{f2}, feature super-resolution \cite{k3} and fuzzy semantic concepts \cite{f32}.

FER in the wild is far more difficult than that in the lab. Some well-performed methods on lab datasets cannot obtain high accuracy on wild datasets. For example, multiscale graph convolutional network \cite{f15} achieved $98.68\%$ accuracy on CK+ but only $86.95\%$ on RAF-DB; FER-VT \cite{f6} achieved $98.68\%$ on CK+ but $84.31\%$ on RAF-DB; TDGAN \cite{f29} achieved $97.53\%$ on CK+ but $81.91\%$ on RAF-DB; as well as OAENet \cite{f5} achieved $98.5\%$ on CK+ but $86.5\%$ on RAF-DB and $58.7\%$ on AffectNet-8. This illustrates that FER in the wild environment requires a more effective model than that in the lab environment.
Xia et al. \cite{f16} proposed ADC-Net by joining channel attention, redesigning the reconstruction
module, and optimizing the loss function based on DCL, which focused on the feature of key local subregions and enabled to improve the accuracy in the wild.
Chen et al. \cite{f34} proposed a novel residual learning module for multi-task learning on facial landmark localization and FER, which can enhance the features and accuracy of both two tasks.
Combining coarse-fine (C-F) labels strategy and distillation, Li et al. \cite{f35} proposed a knowledgeable teacher network (KTN) integrating the outputs of coarse and fine streams, learning expression representations from easy to difficult, which achieved superior performance on several wild datasets.
Applying local attention mechanism, Zhao et al. \cite{f25} proposed a global multi-scale and local attention network (MA-Net), which was capable of acquiring robust both global and local features and can address the issues both of occlusion and pose variation well.
Combining three components that Multi-head Self-Attention Dropping (MSAD), ViT-FER and Multi-Attention Dropping (MAD), Xue et al. \cite{f17} proposed TransFER model which located more diverse relation-aware local representations and achieved higher accuracy on several wild datasets than previous research.

\subsection{Feature Fusion and Ensemble Strategies}
In addition to the strategies reviewed in Section \ref{sec2.1}, feature fusion and ensemble strategies have also been applied to FER.

Firstly, we review some work using feature fusion strategies in FER. Ghazouani \cite{f36} proposed GP-FER with a genetic programming (GP) based framework for selection and fusion of features allowing to combine the hybrid facial features. In GP-FER \cite{f36} based on the genetic algorithm, the most prominent features were selected and fused differently for each pair of expressions so as to improve the performance of the FER model.
Lin et al. \cite{f37} proposed a novel and efficient orthogonalization-guided feature fusion network (OGF$^2$Net) for 2D+3D FER using two networks, FE2DNet and FE3DNet, to extract facial features from 2D and 3D faces. The two networks of OGF$^2$Net were separately trained for late feature fusion and fusing the features of 2D and 3D expression can improve the performance of FER of 2D and 3D mutually \cite{f37}.
Sui et al. \cite{f38} proposed a feature fusion network with masks (FFNET-M) which improves the accuracy of FER by fusing the local features in the salient regions of 2D faces and 3D faces.

Then, we review some work using ensemble strategies in FER.
Renda et al. \cite{f39} compared several ensemble strategies to provide some indications for building up effective ensembles of CNNs to improve the performance of FER and concluded that bagging ensured a high ensemble gain, but the overall accuracy was limited by poor-performing base classifiers as well as the classic averaging voting proved to be an appropriate aggregation scheme, achieving accuracy values comparable to or slightly better than the other experimented operators.
Sun et al. \cite{f40} trained four different networks respectively and integrated them to make the final classification of FER by four decision-level ensemble strategies including nearest neighbor, SVM, probability maximum and probability averaging, which improved the final accuracy of FER.
Benamara et al. \cite{f20} applied the label smoothing technique to deal with the miss-labeled data and utilized the probability distribution-based CNN ensembling method to the FER accuracy.
Chirra et al. \cite{f21} designed a multi-block deep convolutional neural networks (DCNN) model to extract discriminative features, as well as proposed DCNN-SVM based on bagging ensemble with SVM and DCNN-VC based on voting technique, where the majority voting technique was utilized to vote between SVM, RF, and LR classifiers to make better predictions of FER.
Wadhawan et al. \cite{e3} proposed a Part-based Ensemble Transfer Learning network that modeled how humans recognize facial expressions by correlating the spatial orientation pattern of the facial features with a specific expression.
Choi et al. \cite{e4} combining deep convolutional neural networks
with stochastic ensemble weight optimization and proposed DCNN ensemble classifier.
Karnati et al. \cite{e5} proposed a texture-based feature-level ensemble parallel network for FER (FLEPNet)  to solve the aforementioned problems, which uses multi-scale convolutional and multi-scale residual block-based DCNN as
building blocks.

From the above literature review,
feature fusion in FER mainly focused on the fusion within the same network or between 2D and 3D faces;
the existing ensemble strategies in FER, leveraging SVM, voting and probability maximum, mainly integrated several networks with different structures and mainly focused on the ensemble ranking first in the classifiers. Additionally, there is no research using both feature fusion and ensemble strategies simultaneously.
Different from previous studies, this paper applies both internal feature fusion for a single model and feature fusion among multiple networks, as well as the ensemble strategy that Top Two Voting (T2V). The T2V integrates networks with the same structures and considers more ranking of classifiers' output. Finally, not only does the single model FAML achieve a high-level performance, but also the ensemble model FAML-FGA-T2V outperforms the state-of-the-art methods on several challenging unbalanced FER datasets in the wild.

\begin{figure}[ht!]
            \centering
            \includegraphics[width=0.99\columnwidth]{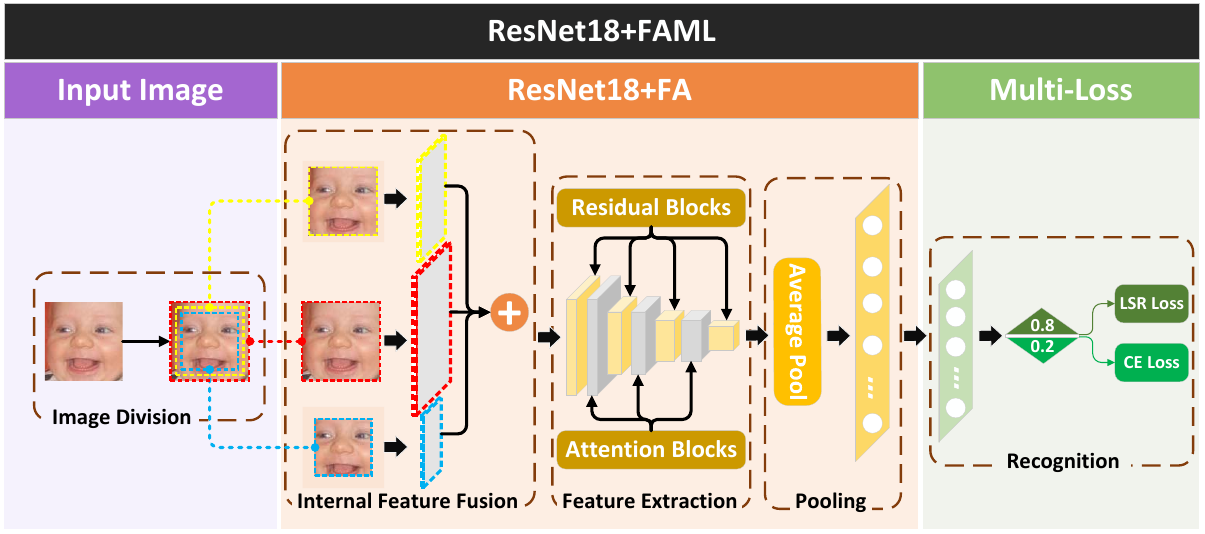}
            \caption{The architecture of the proposed ResNet18+FAML.}
            \label{fig3}
\end{figure}
\begin{figure}[ht!]
  \centering
  \includegraphics[width=\linewidth]{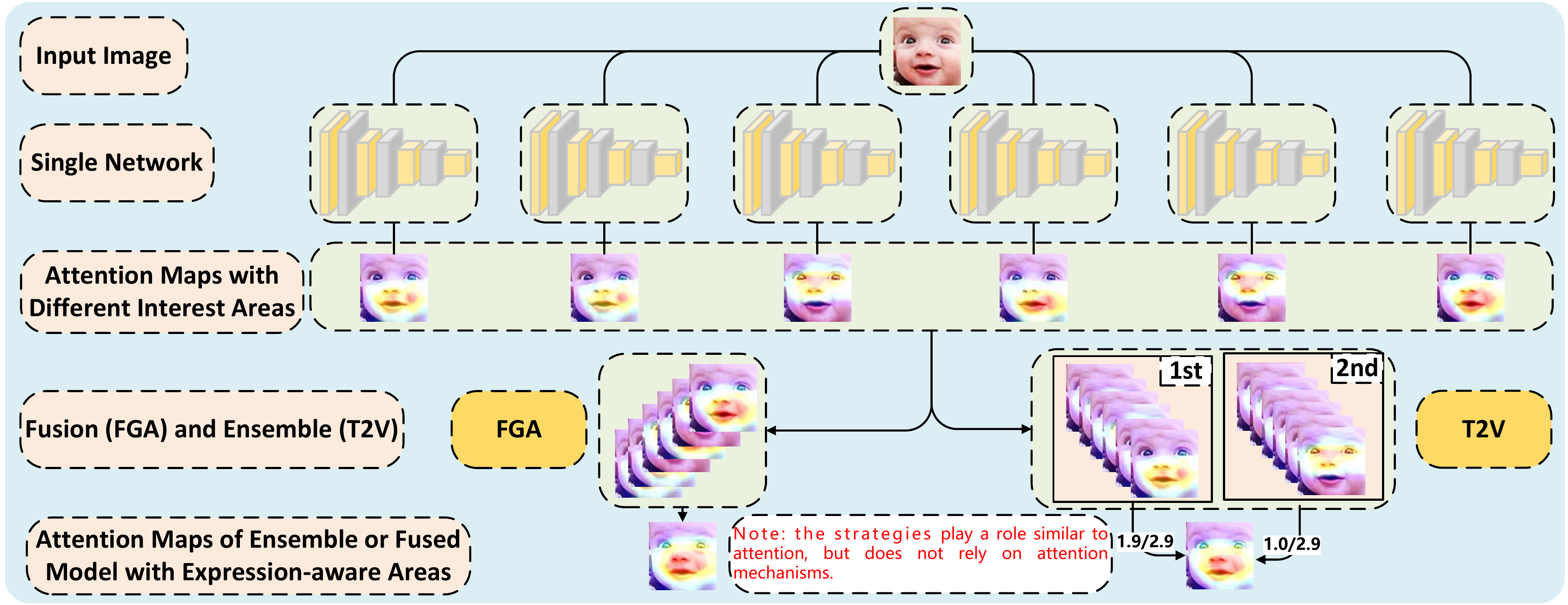}
  \caption{The diagrams of ensemble and fusion strategies for facial expression recognition: integrating the interest areas of multiple networks to focus on main expression-aware areas for higher accuracy. }
  \label{fig0}
\end{figure}

\begin{figure}[ht!]
            \centering
            \includegraphics[width=0.99\columnwidth]{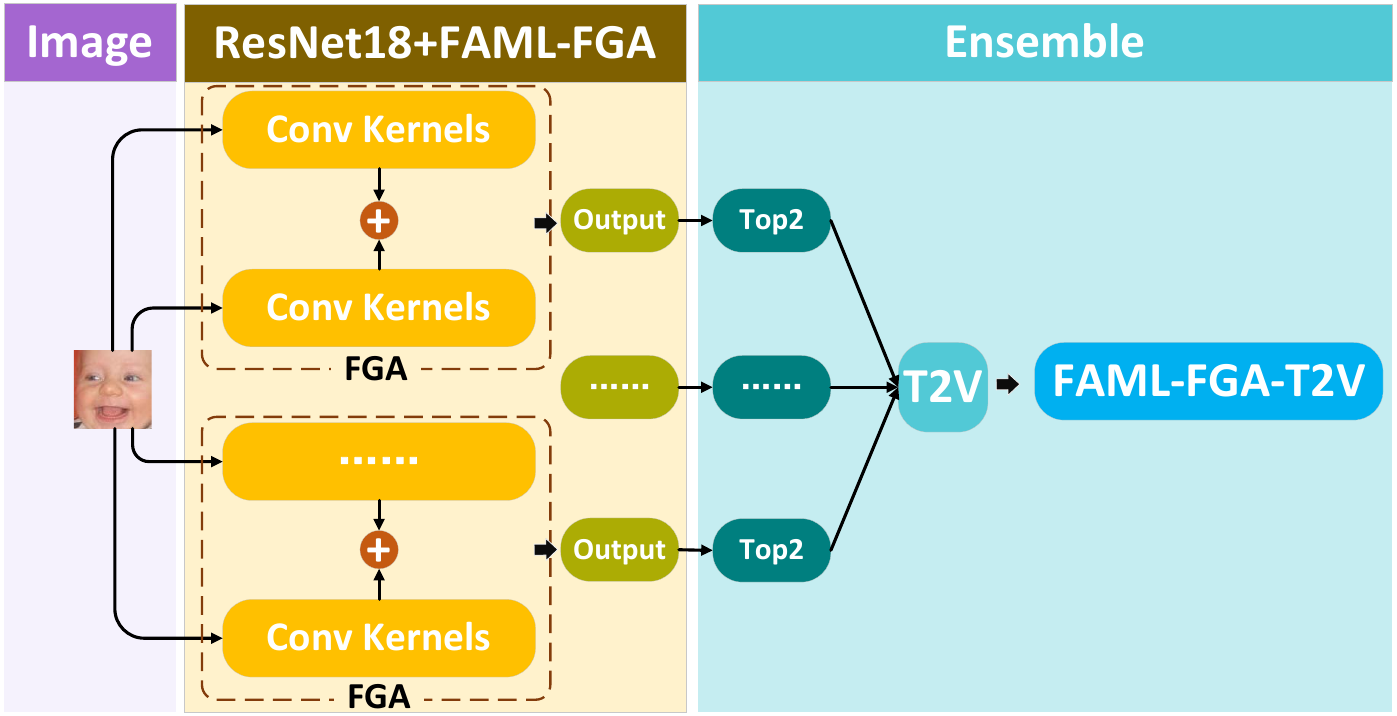}
            \caption{The frameworks of the proposed R18+FAML-FGA-T2V.}
            \label{fig6}
\end{figure}

\begin{figure*}[ht!]
            \centering
            \includegraphics[width=1.49\columnwidth]{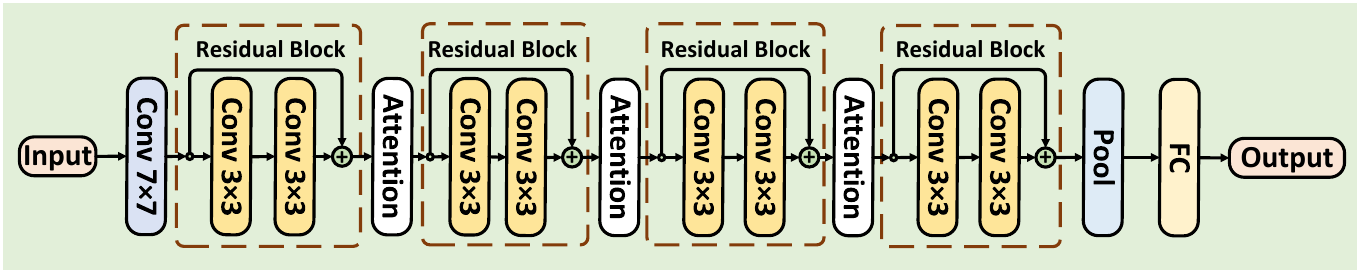}
            \caption{The architecture of ResNet18 with three Attention Blocks.}
            \label{fig4}
\end{figure*}

\section{Network and Methodologies}\label{sec3}
Simply but efficiently, our proposed structure of R18-FAML is shown as Fig. \ref{fig3} and the diagrams of ensemble and fusion strategies for FER to integrate interest areas of multiple networks to focus on main expression-aware areas is shown as Fig. \ref{fig0}.
The frameworks of our proposed ensemble model R18+FAML-FGA-T2V are shown as Fig. \ref{fig6}.  As mentioned above, due to illumination, occlusion, large similarities and unbalanced data amount between categories, it is highly desired to build a more effective model to address FER in the wild. To achieve that, we comprehensively design several significant components: the network architecture with training strategies, feature fusion among multiple networks, and ensemble strategy.

\subsection{Network Architecture and Training Strategies}%可以考虑合并网络和训练策略
As shown in Fig. \ref{fig4}, we use ResNet18 \cite{f12} with three attention blocks as feature extraction module.

ResNet18 \cite{f12} introduced the residual structure can solve the gradient disappearance problem to some extent, but it still faces serious overfitting in the wild FER without the application of other strategies.
One of the strategies to solve that is the attention mechanism.
ResNet18 has four residual blocks (res\_blocks).
After the first three res\_blocks, we respectively insert an attention block (att\_block). In the att\_blocks, we use ViT \cite{f41} to enhance the weights of the features in the local region.

On the basis of this architecture, we apply internal feature fusion to modify the extracted features.
For the convolutional neural network layer of the first layer of ResNet18, we replace it with three convolutional neural networks. As shown in Fig. \ref{fig3}, the two convolution networks are trained by the part of the image and their feature maps (the yellow map and blue map) are fused to the main feature map (the red map) by weighted addition, which can assist in modifying the main convolution layer. This internal fusion feature enhances the feature extraction ability by magnifying the effective feature channels. FA sets three (arbitrarily changeable in practice) areas of different sizes based on the proportion of facial features, which increases the probability of interest areas to affect the parameters updates and also maintains overall information.

To better train the network, we consider using multiple losses and variable learning rates as Fig. \ref{fig3}.
For multiple losses, we calculate loss randomly by two loss functions with the given probabilities: Label Smoothing Regularization (LSR) loss and Cross Entropy (CE) loss.
For the cross entropy loss, the label vector $p^{CE}_i$ is Equation (\ref{eq01}) and loss is Equation (\ref{eq02}).
\begin{equation}\label{eq01}
p^{CE}_i = \left\{
\begin{aligned}
1,&\quad i=y\\
0,&\quad i\ne y\\
\end{aligned}
\right.,
\end{equation}
\begin{equation}\label{eq02}
\mathfrak{L}_{CE} \left(p,q \right) =-\sum\limits^C_{i}p^{CE}_i\log{q_i}
\end{equation}
where $C$ is the number of classes and $q\left(i\right)$ is the probability of the $i$-th class after softmax processing as:
\begin{equation}
q_i = \frac{exp\left(z_i\right)}{\sum^{C}_{j=1}exp\left(z_j\right)}
\end{equation}
assuming the output vector of network as $Out = \left\langle z_1, z_2, \dots, z_C \right\rangle$.

Label smoothing is a regularization strategy, which reduces the weight of the real sample label category when calculating the loss of networks, and suppresses overfitting. In LSR, we set up the new label vector as

\begin{equation}
P^{LSR}_i = \left\{
\begin{aligned}
\left(1 - \varepsilon \right),&\quad i=y\\
\frac{\varepsilon}{C-1},&\quad i\ne y\\
\end{aligned}
\right.,
\end{equation}
where $\varepsilon \in \left[0,1\right]$, then the LSR is calculated as:
\begin{equation}
\mathfrak{L}_{LSR} \left(p,q \right) =-\sum\limits^C_{i=1}P^{LSR}_i\log{q_i}
\end{equation}

ML randomly selects a loss in each batch, e.g., 20\% probability of choosing CE and 80\% of LSR. It's different from existing methods (e.g., 0.2CE+0.8LSR). ML provides two search paths for updates, thereby multi-path search. But, 0.2CE+0.8LSR corresponds to single-path. In theory, the local optimum of multi-path search satisfies the convergence conditions of all paths simultaneously, hence usually better than that of single-path.

For learning rate, we set up two learning rate strategies as $0.0003$ with Cosine Annealing scheduler ($Lr_1$ ) and $0.00001$ with StepLR scheduler ($Lr_2$ ). Initially, we use $Lr_1$, and we switch the learning rate strategies from $Lr_1$ to $Lr_2$ when the loss of the network achieves a stable state.

With these strategies, we can obtain our basic model as R18+FAML using Feature fusion and Attention mechanism with Multiple Losses shown as Fig. \ref{fig3}.

\subsection{Feature Fusion between Models based on GA}
Although FAML can achieve excellent performance in FER, we still consider continuing to improve the performance of the model. Therefore, we proposed feature fusion among multiple networks based on the genetic algorithm (FGA).

The training process of the deep neural network can be regarded as the parameter search. In CNN, the weights of convolution kernels are actually the feature extractor. We can assuming a convolution layer has $m$ out-channels and $n$ in-channels and each kernel size is $(s_1,s_2)$, which means there are $m$ filters and each filter has $n$ convolution kernels. The parameters matrix of the $i$-th kernel $k_{ji}$ in the $j$-th filter $f_j$ can be set as $W_{ji}=\left\{w_{lh}\right\}_{s_1\times s_2}$. Additionally, we can set the input of this convolution layer as $D = \left\langle D_1, D_2, \dots , D_n \right\rangle$. According to the definition of convolution operation in CNN, the output of the input data $D_i$ after the operation of the kernel $k_{ji}$ is $O_{ji}= D_i \otimes W_{ji}$ where $\otimes$ means convolution operation of convolution layer. Assuming the operation of merging the output results of convolution kernels in the same filter into feature maps is $\textbf{H}$, we can obtain the matrix corresponding to filter $f_j$ as:
\begin{equation}\label{eq2}
F_j = \mathop \textbf{H}\limits_{i = 1}^n \left( {{O_{ji}}} \right)=\mathop \textbf{H}\limits_{i = 1}^n \left( {{D_i \otimes W_{ji}}} \right)
\end{equation}

The matrix $F_j$ is actually the feature map extracted by the convolution neural network, which will participate in further feature extraction or classification.
In the test of ensemble strategies, we get a conclusion that linearly adding the output of multiple networks with the same structure can improve the overall accuracy. This enlightens us to add the features extracted from the convolution layer in multiple networks.
For $M$ networks with the same structure, we can set the feature of the $j$-th filter in the $l$-th network $N_l$ as $F^{l}_{j}$ where $F^{l}_{j} = \mathop \textbf{H}\limits_{i = 1}^n \left( {{D^{l}_i \otimes W^{l}_{ji}}} \right)$ referring to Equation (\ref{eq2}). Fusing the feature maps of these $M$ networks by linear average can obtain a new feature map as:
\begin{equation}\label{eq3}
F^{(f)}_{j} =  \frac{1}{M}\sum\limits_{l = 1}^M {F_j^l}  = \frac{1}{M}\sum\limits_{l = 1}^M {\mathop \textbf{H}\limits_{i = 1}^n \left( {{D^l_i} \otimes W_{ji}^l} \right)}
\end{equation}

The operation of Equation (\ref{eq3}) is a way to forcibly change the feature map of the convolution layer fusing features extracted by multiple networks. The fused network cannot directly participate in the classification, which requires re-training. However, if the network is trained directly on the basis of Equation (\ref{eq3}), the update of each network will affect mutually, which will cause the overall overfitting of the network after the fusion. Therefore, we consider whether we can directly modify the kernel parameters of the convolution layer without influencing the performance of feature fusion. The special operation property of the convolution network makes it possible. When the operation $\textbf{H}$ is addition, Equation (\ref{eq3}) can be rewritten as
\begin{equation}\label{eq4}
F^{(f)}_{j}  = \frac{1}{M}\sum\limits_{l = 1}^M {\sum\limits_{i = 1}^n \left( {{D^l_i} \otimes W_{ji}^l} \right)}
\end{equation}

For the networks with the same structure, their abilities of feature extraction are approximate. Thus, it is approximative that $D^{l_1}_i \approx D^{l_2}_i \approx D_i$ where $1\le l_1 \le l_2\le M$. Substitution it into Equation (\ref{eq4}) and consideration of the Distribution Law of convolution operation can obtain
\begin{equation}\label{eq5}
F^{(f)}_{j}  \approx \sum\limits_{i = 1}^n {\left( {D_i \otimes \frac{1}{M} \sum\limits_{l = 1}^M {W_{ji}^l} } \right)}.
\end{equation}
$\frac{1}{M} \sum_{l = 1}^M {W_{ji}^l} $ in Equation (\ref{eq5}) reveals that the direct average of the weights of convolution kernels in multiple networks can realize feature fusion among networks, which also allows each network to be trained independently.

Based on Equation (\ref{eq5}), we forcibly update the convolution layer parameters in our designed main network i.e. FAML, when the loss change is relatively stable after training.
In multiple networks, only part convolution kernels of some networks have a great impact on the feature map. Therefore, we eliminate some convolution kernels with poor features and retain the useful part through the optimization method of the genetic algorithm resulting in well-performed networks. Finally, we obtain the algorithm of GA-based feature fusion among networks (FGA) as Algorithm \ref{algo1}, which improves the accuracy of the single model. Commonly, `well-trained' is estimated by accuracy or loss in training or testing datasets. Our used criterion is fluctuation ranges of accuracy and loss in training datasets within a given number of successive epochs are less than threshold. The trigger for the learning rate decrease in the paper is analogous.

\begin{algorithm}\label{algo1}

Input multi networks after well training

\For{$i$ in number of fused networks}
{
Calculate the probability of each network to be selected according to the loss in the training process

Randomly and preferentially select a given number of the networks

Obtain the weights of each convolution layer in the selected networks

Add weights of selected networks in each channel as Equation (\ref{eq5})

Obtain the $i$-th fused networks}

Initialize multi new networks

Train the new initialized networks and fused networks

Take the well-trained networks as the input for the next FGA

\caption{FGA: GA-based feature fusion among networks}
\end{algorithm}

\subsection{Ensemble Strategy: T2V}
With the combination of FAML and FGA, FAML-FGA can achieve an appreciable performance in FER. On this basis, we continue to explore another strategy to improve the accuracy of FER.

In a large number of experiments, we observe that many samples with the wrong prediction ranked second in the classification output.
Since most networks have similar conclusions, we only choose three networks including the RI18 (ResNet18 pre-trained on ImageNet), R18 (ResNet18 pre-trained on Ms-Celeb-1M \cite{f43}) and R18+FAML as the examples on RAF-DB dataset to illustrate this observation, whose probability distributions of the labeled category ranked in the output are shown as Table \ref{table110}.
From Table \ref{table110}, the probability that the RI18, R18 and R18+FAML identify the correct label as the first-ranking are $85.53\%$, $88.63\%$ and $90.32\%$ respectively, as well as that as the second-ranking are $9.35\%$, $7.59\%$ and $6.19\%$ respectively. That means the probabilities that the correct label appears in the top two classes of these networks' outputs are $94.88\%$, $96.22\%$ and $96.51\%$.
These examples indicate that if the second-ranking category can be taken into account, it is possible to improve the accuracy of FER. Aiming at this target, we propose an ensemble strategy, i.e. T2V (Top Two Voting).

\begin{table}[ht!]

\begin{center}
\caption{The probability distribution (\%) of the ranking of labeled categories in the output of several networks on RAF-DB datasets. }
\label{table110}
\setlength{\tabcolsep}{3pt}
%\begin{tabular}{p{55pt}p{55pt}p{190pt}p{100pt}}
\begin{tabular}{c|cccccccc}
\hline
\multirow{2}{*}{Networks} & \multicolumn{7}{c}{Rank}\\
\cline{2-8}
& 1 &  2 & 3 &  4&  5 &  6 &  7\\
\hline
RI18  & 85.33& 9.35& 2.77& 1.37& 0.62& 0.29& 0.26\\
R18 & 88.63& 7.59& 2.25& 0.68& 0.52& 0.26& 0.07\\
R18 + FAML& 90.32& 6.19& 1.96& 0.78& 0.29& 0.33& 0.13\\
\hline

\end{tabular}
\end{center}
\end{table}

\begin{table}[ht!]
\begin{small}
\begin{center}
\caption{The evaluation (\%) of attention blocks with internal feature fusion (FA), loss function, genetic-based feature fusion among models (FGA), and ensemble strategies on RAF-DB, AffectNet-8 and  AffectNet-7. }
\label{table10}
\setlength{\tabcolsep}{1pt}
\begin{tabular}{cccccccc}
\hline
 Models & FA &  Loss & FGA & T2V& RAF-DB & AffectNet-8 & AffectNet-7\\
\hline
R18  & & & & & $88.62$ & 52.16 & 58.17\\
R18+FA    & \Checkmark & & & & $88.72$ & 52.54 & 59.77\\
FAML    &\Checkmark & LSR+CE & & & $90.32$ & \textbf{62.17} & 65.83\\
FAML-FGA    &\Checkmark & LSR+CE & \Checkmark & & $90.51$ & \textbf{62.29} & 65.89\\

FAML-T2V    &\Checkmark & LSR+CE &  & \Checkmark & $\textbf{91.46}$ & \textbf{63.04} & \textbf{66.51}\\
FAML-FGA-T2V    &\Checkmark & LSR+CE & \Checkmark & \Checkmark & $\textbf{91.59}$ & \textbf{63.27} & \textbf{66.63}\\
\hline

\end{tabular}
\end{center}
\end{small}
\end{table}

\begin{table*}[ht!]

\begin{center}
\caption{The outputs of six single networks (S\_Ns) and their ensemble model on an ``Anger" image in RAF-DB dataset. }
\label{table112}
\setlength{\tabcolsep}{1pt}

\begin{tabular}{c|ccc}
\hline
Image  &Model & Output & Class \\
\hline
\multirow{8}{*}{ \includegraphics[width=0.13\columnwidth]{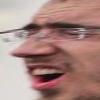}}
&S\_N 1 & $\left[0.295,	0.138,	-1.331,	0.917,	-1.429,	-0.502,	-0.712\right]$ & FE   \\
&S\_N 2 & $\left[0.312,	0.058,	-1.383,	0.888,	-1.426,	-0.547,	-0.495\right]$ & FE \\
&S\_N 3 & $\left[0.187,	-0.119,	-0.835,	1.715,	-1.240,	-0.118,	-1.127\right]$ & FE \\
&S\_N 4 & $\left[1.518,	-0.168,	-1.641,	0.489,	-1.482,	-0.985,	-0.491\right]$ & AN \\
&S\_N 5 & $\left[0.798,	-0.210,	-1.823,	0.753,	-1.308,	-0.582,	-1.493\right]$ & AN \\
&S\_N 6 & $\left[0.403,	-0.433,	-1.392,	-0.637,	-0.493,	-0.131,	-1.072\right]$ & AN \\
\cline{2-4}
%&NOI & $\left[1.7425,	0.7925,	0.2242,	\textbf{1.9689},	0.3011,	0.5895,	0.3813\right]$ & FE \\
&T2V & $\left[\textbf{8.7},	0,	0,	7.7,	0,	1,	0\right]$ & AN \\

\hline

\end{tabular}
\end{center}
\end{table*}

In T2V (Top Two Voting), the top two categories of each network output are counted with a certain weight to participate in voting. Assuming the index vector of $Out^l$ of the $l$-th network after descending sorting is $S^l = \left\langle s^l_1, s^l_2, \dots, s^l_C \right\rangle$, we can obtain a label vector $V^l =\left\langle v^l_1, v^l_2, \dots, v^l_C\right\rangle$ as
\begin{equation}\label{eqv}
v^l_i = \left\{
\begin{aligned}
\alpha, &\quad s^l_i = 0;\\
\beta , &\quad s^l_i = 1;\\
0, &\quad s^l_i > 1;\\
\end{aligned}
\right.
\end{equation}
where $\alpha$ and $\beta$ are the weights of the first and the second-ranked categories respectively.
In order to calculate quickly in the program, we can use Equation (\ref{eq90}) to replace Equation (\ref{eqv}) to obtain $V^l$:
\begin{equation}\label{eq90}
\begin{aligned}
V^l &= \left(\alpha - 2\beta\right)\cdot{\rm maximum}\left( \left(S^l + 2 - C\right),0\right) \\
    &+ \beta\cdot {\rm maximum}\left(\left(S^l +3 - C\right),0\right)
\end{aligned}
\end{equation}
Then the T2V output is
\begin{equation}\label{eq9}
Out^{T2V} = \sum\limits^{M}_{l=1} V^l
\end{equation}

In the training process, the random flipping of images and the multiple loss functions can both increase the difference of networks so as to ensure the effectiveness of T2V. Additionally, $Out^{T2V}$ is only utilized for classification without participating in training.
Obviously, T2V eliminates the inconsistency of the output order of each network and doesn't need normalization. T2V simultaneously puts the second-ranking category into consideration better. 

In FER, each network can provide some effective feature extractors to address a subset of challenges (illumination, occlusion, etc.). FGA and T2V actually hold effective extractors to enrich the interest areas to increase their impact, hence they can enhance the accuracy of FER, seen as Fig. \ref{fig0}. Although the structures, pre-trained weights etc. of multi networks are the same, the randomness, e.g., random loss, initial weights of FC, data-flipping, takes networks to different convergence mapping regions. Therefore, multi networks well-trained will provide different effective feature extractors. Thus, our proposed FGA and T2V can performed their effect to enhance the accuracy of classifications.

Combining FAML, FGA and T2V, we finally obtain an ensemble FER model, i.e. FAML-FGA-T2V, whose frameworks are shown in Fig. \ref{fig6}.

\section{Experiments}\label{sec4}
\subsection{Experiments Design and Datasets}
In this section, we verify our proposed models from several aspects: attention blocks with internal feature fusion (FA), loss function, feature fusion among networks, and ensemble strategy.
In order to verify the effectiveness of these aspects, we set up several groups of ablation experiments;
present the results of our proposed models compared with existing methods on several unbalanced FER datasets in the wild including RAF-DB \cite{f7}, AffectNet-8 \cite{f8}, and AffectNet-7 \cite{f8}; study the universality of our proposed ensemble strategy; finally, present the attention visualization of proposed models on RAF-DB dataset. The descriptions of the datasets used in experiments are as follows.

\begin{itemize}[leftmargin=*, topsep = 0em, partopsep=0em, itemsep=0em, parsep = 0em]
\item RAF-DB \cite{f7}: is a large-scale affective face database with large diversities and rich annotations in the real world, which contains 30,000 facial images annotated with basic (7 classes of basic emotions) or compound (12 classes of compound emotions) expressions by 40 annotators. In our experiment, we use the 7 basic emotions: Anger (AN), Disgust (DI), Fear (FE), Happiness (HA), Neutral (NE), Sadness (SA), Surprise (SU), which includes 12271 images as training data and 3068 images as test data.
\item AffectNet \cite{f8}: is currently the largest facial expression dataset in the wild, which contains one million images collected from the Internet by querying expression-related keywords in three search engines, in which 450,000 images are manually annotated with 11 expression categories. In our experiments, we evaluate our models on both AffectNet-8 and AffectNet-7. AffectNet-7 contains six basic expressions (Anger, Disgust, Fear, Happiness, Sadness, Surprise) and Neutral expressions with 283901 images as training data and 3500 images as test data. And AffectNet-8, with 287568 training images and 4000 test images, has the addition of Contempt (CO) expressions based on AffectNet-7.
\end{itemize}

All images are aligned and resized to $224\times 224$ pixels. In each batch of the training process, we flip the images horizontally with a 50\% probability. To balance the class distribution, we over-sample the training data by adding weight to the CE loss function.
The parameter $\varepsilon$ of LSR loss is set up as $0.1$. The networks are trained with the Adam optimizer to minimize the corresponding loss function with the mini-batch size of 128 for RAF-DB dataset and 256 for AffectNet dataset.
For the loss function, we set 80\% probability of LSR and 20\% of CE on RAF-DB, as well as 20\% probability of LSR and 80\% of WCE (weighted cross entropy) on AffectNet.
The above setting is because the categories proportions of the test set and training set on RAF-DB are approximate, while that of AffectNet are significantly distinct.
For the parameters of T2V, we set up $\alpha = 1.9$ and $\beta = 1.0$, which implies the score of two second-rankings will exceed that of one first-ranking.

The experiments are launched in a distributed system with multi GPU servers with configurations as follows.
\begin{itemize}[leftmargin=*, topsep = 0em, partopsep=0em, itemsep=0em, parsep = 0em]
\item CPU: Intel i9 10850K, 3.6GHz, 10 cores;
\item RAM: LPX 64GB DDR4 3200;
\item GPU: NVIDIA TESLA V100 32GB;
\item Operation System: Ubuntu 18.04;
\item Deep learning framework: Pytorch 1.7.1.
\end{itemize}

\subsection{Ablation Studies}
To evaluate the performance of our proposed strategies including FAML, FGA and T2V, we carry out the ablation studies to investigate the effects of internal feature fusion with attention blocks (FA), loss function, feature fusion among networks, and ensemble strategy.  Then, the accuracies of the models on RAF-DB, AffectNet-8 and AffectNet-7 are shown as Table \ref{table10}, where R18 means the ResNet18 pre-trained on Ms-Celeb-1M \cite{f43}.

\begin{figure}[ht!]
  \centering
  \subfigure[R18 on RAF-DB]{\includegraphics[width=0.49\columnwidth]{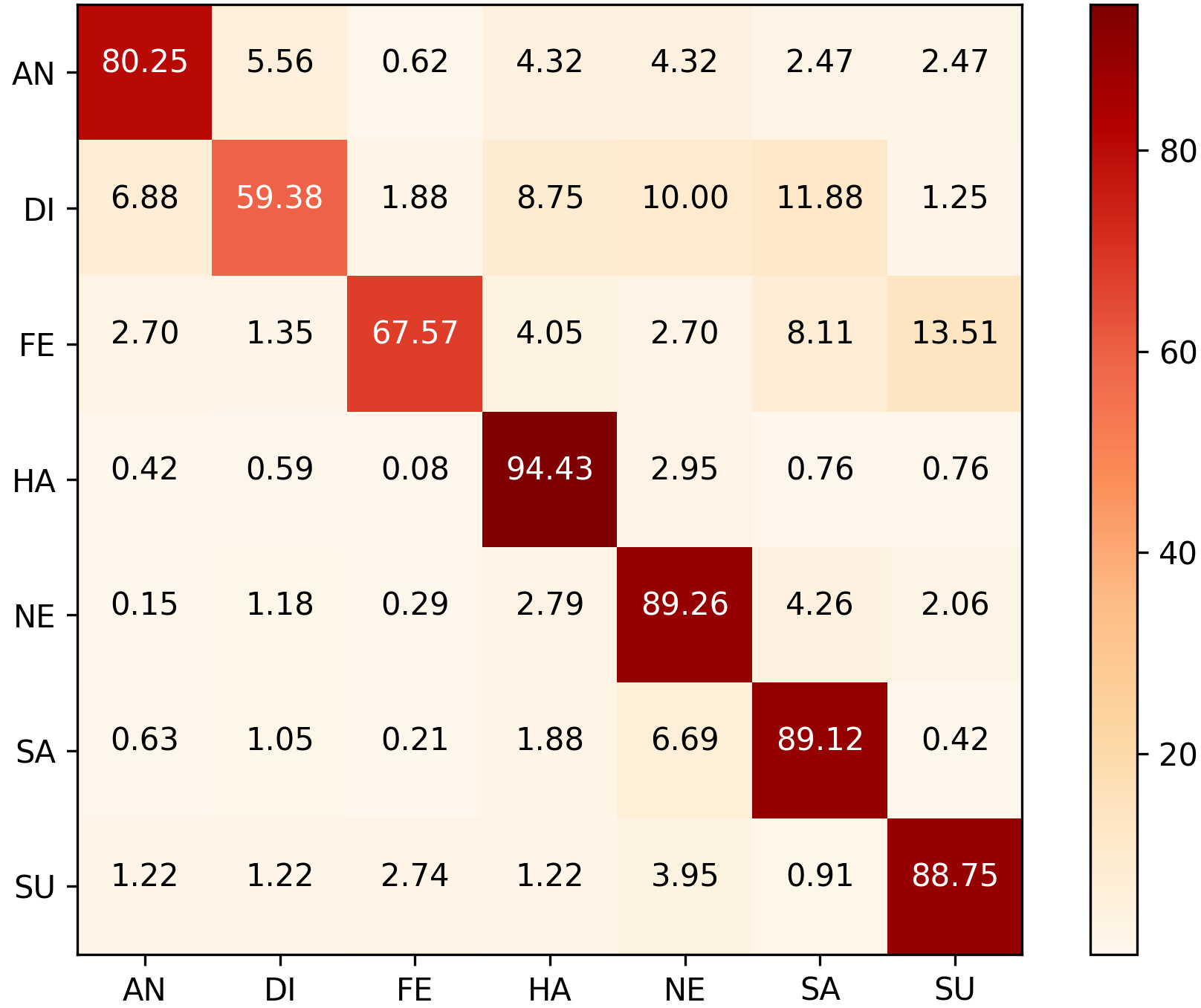}}
  \subfigure[R18+FAML on RAF-DB]{\includegraphics[width=0.49\columnwidth]{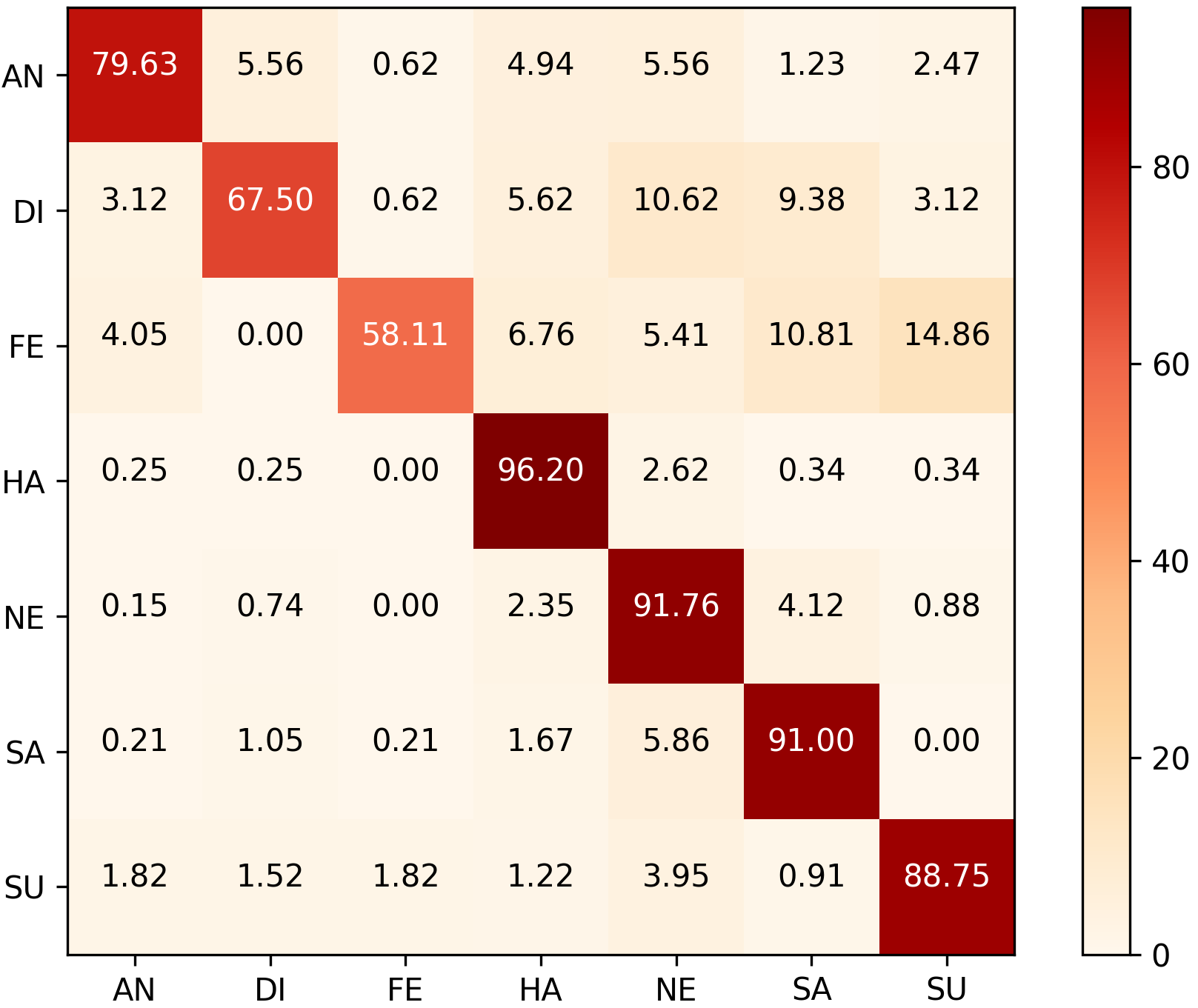}}
  \subfigure[R18+FAML-FGA-T2V on RAF-DB]{\includegraphics[width=0.49\columnwidth]{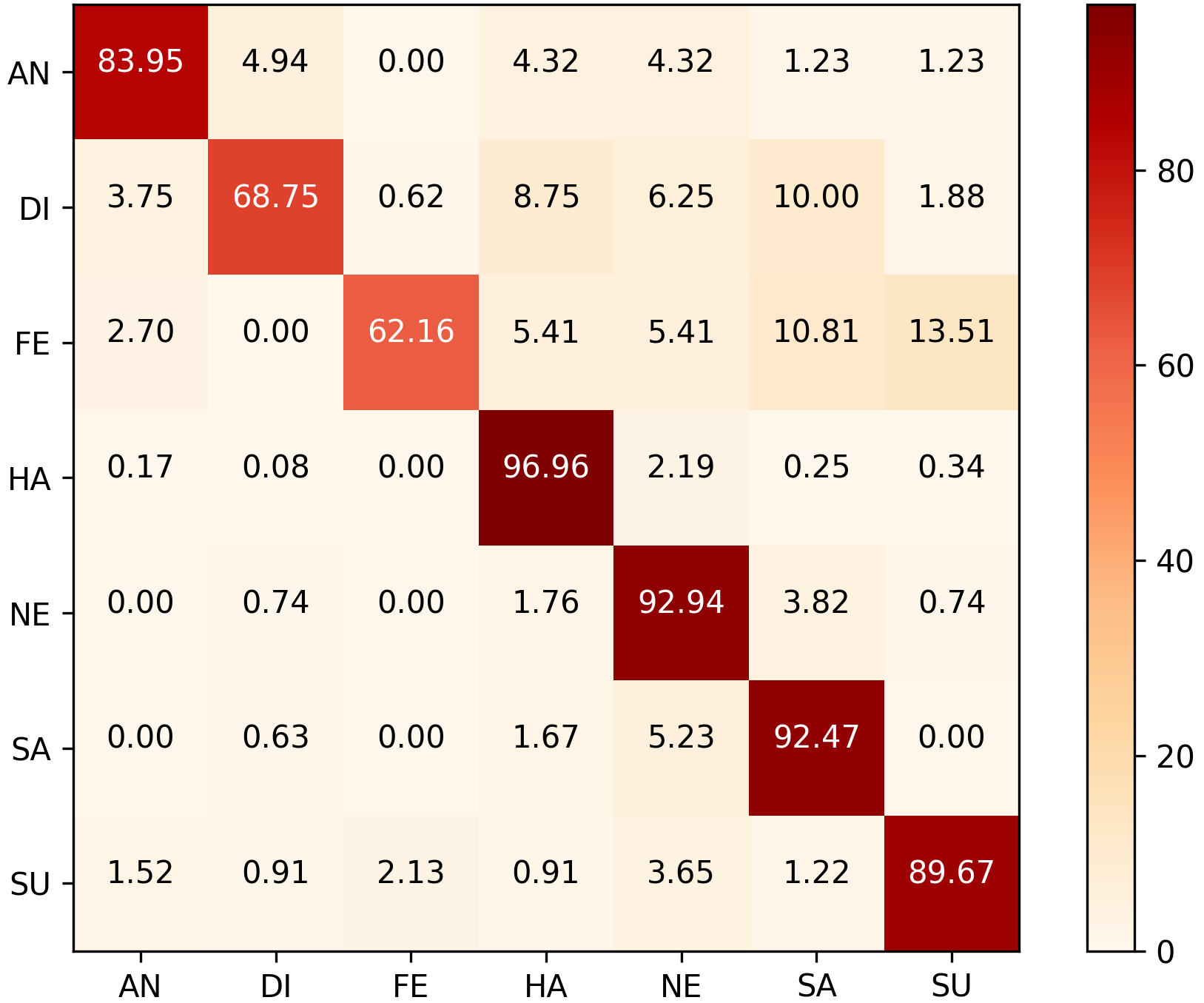}}
  \subfigure[R18 on AffectNet-8]{\includegraphics[width=0.49\columnwidth]{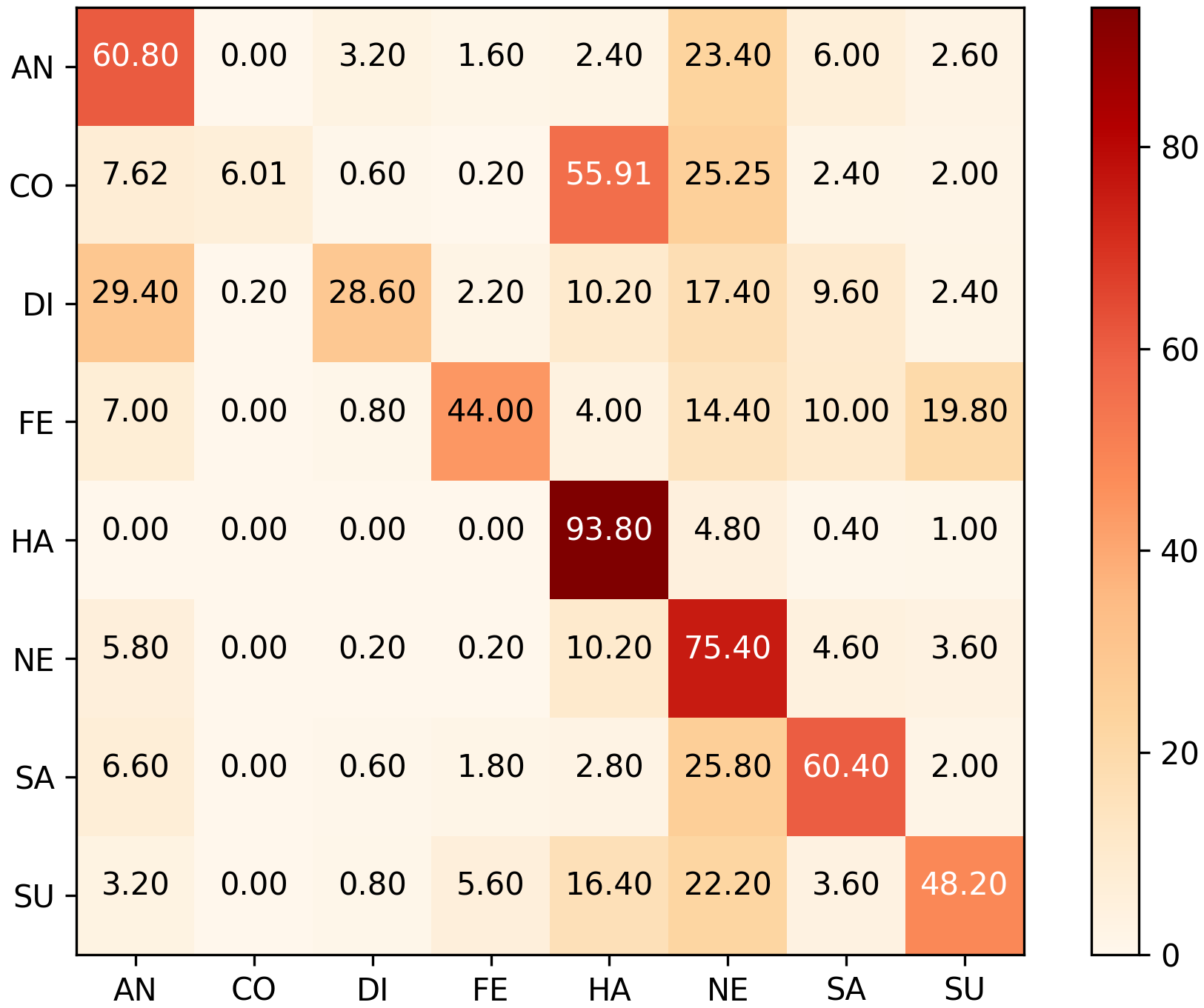}}
  \subfigure[R18+FAML on AffectNet-8]{\includegraphics[width=0.49\columnwidth]{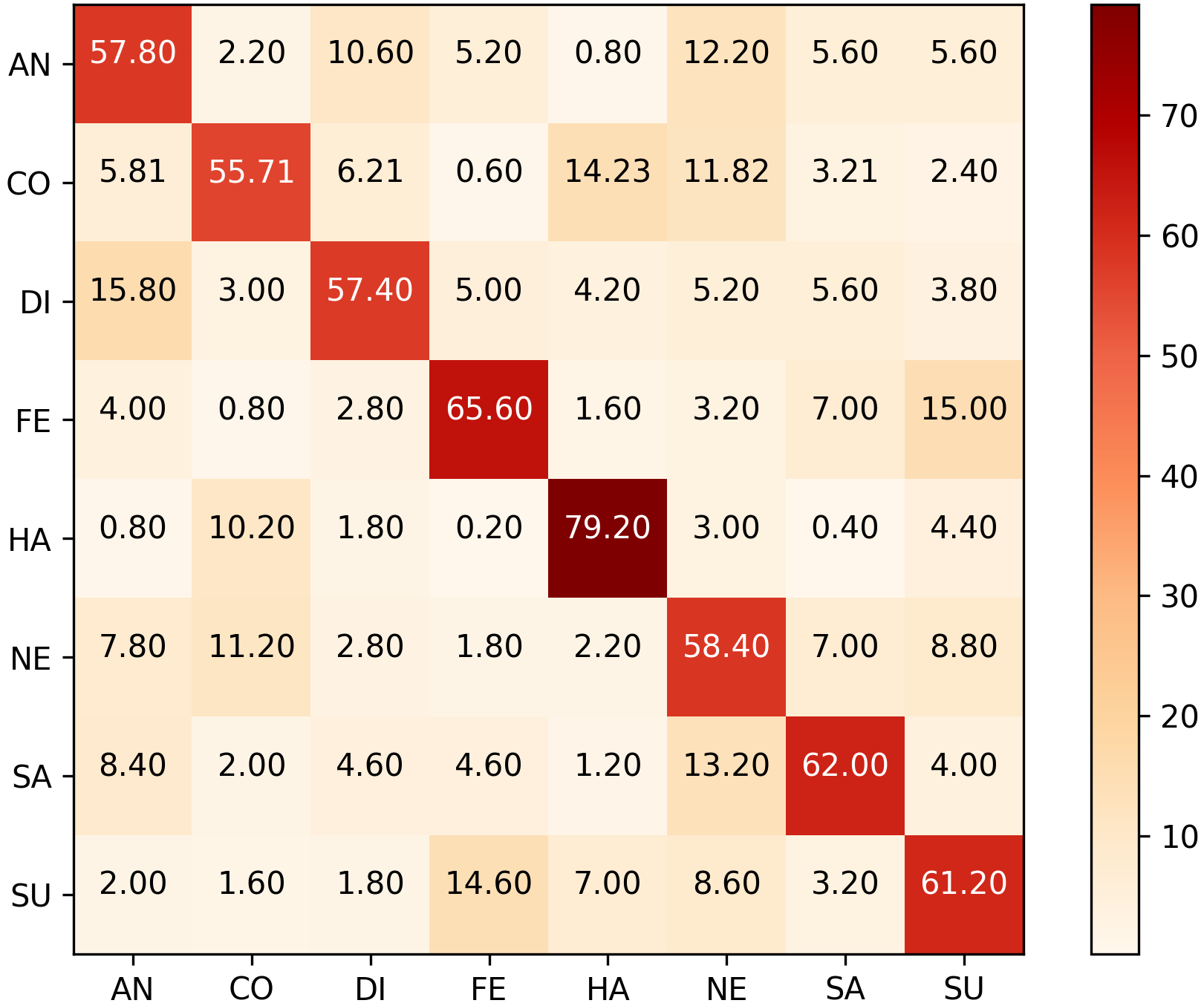}}
  \subfigure[R18+FAML-FGA-T2V on AffectNet-8]{\includegraphics[width=0.49\columnwidth]{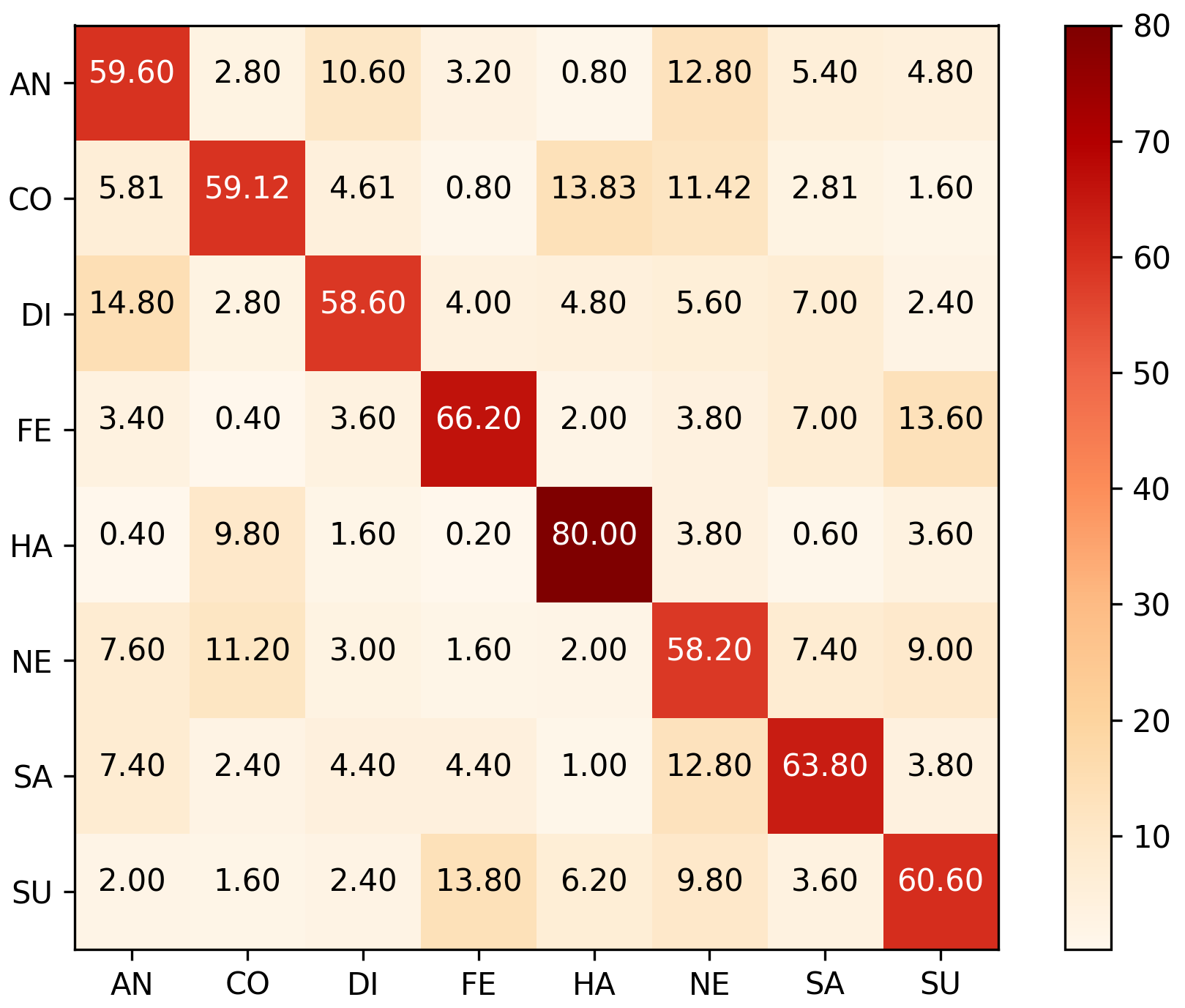}}

  \caption{The confusion matrices (\%) of R18, R18+FAML and R18+FAML-FGA-T2V on RAF-DB and AffectNet-8.}
  \label{fig11}
\end{figure}

For the sake of efficient description and discussion, we use a tuple $\left(a, b, c\right)$ to denote the performance on RAF-DB, AffectNet-8 and  AffectNet-7.
From Table \ref{table10}, the baseline R18 without FA, multi-loss, FGA and ensemble strategy, has achieved $\left(88.62\%, 52.16\%, 58.17\%\right)$. Compared with the baseline, FA improves the performance by $\left(0.10\%, 0.38\%, 1.60\%\right)$, and the FAML (combining FA and LSR+CE) significantly improves the performance  by $\left(1.70\%, 10.01\%, 7.06\%\right)$. Compared with the R18+FAML, R18+FAML-FGA improves the performance by $\left(0.19\%, 0.12\%, 0.40\%\right)$, which indicates the fusion of convolution kernel parameters of multiple models as $\frac{1}{M} \sum_{l = 1}^M {W_{ji}^l} $ can improve the accuracy of classification slightly.
$89\%\to 90\%$ is much harder than $88\%\to 89\%$ although both with $1\%$ increment.
 $0.1\%$ increment is difficult in FER, necessitating effective strategies, e.g. Above $88\%$ in RAF-DB. This also indicates that the significance of FGA cannot be measured by absolute improvement.
In fact, we changed the optimizer from ``Adam" to ``SGD" with 0.005 learning rate and re-trained the R18+FAML with $90.32\%$ on RAF-DB, then we also obtain a model with an accuracy of $90.51\%$ however relying on a great influence of parameter adjustment. This phenomenon also reveals that multi-loss still has potential.
For the ensemble strategy, R18+FAML-FGA-T2V increases the accuracy by $\left(1.08\%, 0.98\%, 0.94\%\right)$  compared with R18+FAML-FGA; FAML-T2V increases the accuracy by $\left(1.14\%, 0.87\%, 1.28\%\right)$  compared with R18+FAML. Utilization of ensemble strategy T2V enables the FAML-FGA-T2V to achieve the state-of-the-art performance $\left(91.59\%, 63.27\%, 66.63\%\right)$.

The comparison between R18+FAML-FGA and R18+FAML-T2V shows that T2V strategy has a better effect to improve the performance of models, which is because T2V directly scores the ranking of classification ignoring the quantitative influence of output value. Deep networks usually use the category ranking first in the output as the classification result, and the specific values of the output from the different images may be quite different. For example, the image labeled as an ``Anger" image from RAF-DB, and its outputs for six well-trained R18+FAML models and their ensemble model are respectively as Table \ref{table112}.
From Table \ref{table112}, T2V gets the correct class ``Anger" with the output  $[8.7,	0,	0,	7.7,	0,	1,	0]$.

To evaluate the overall performance, we apply the confusion matrices of the models in Table \ref{table10}. Since the effects of other models are similar, we only present the confusion matrices of R18, R18+FAML and R18+FAML-FGA-T2V on RAF-DB and AffectNet-8 datasets in Fig. \ref{fig11}. From Fig. \ref{fig11}, ``Disgust" and ``Fear" are the most difficult expressions for R18 with the lowest recognition accuracies on RAF-DB, while ``Contempt", ``Disgust" and ``Fear" are the most difficult expressions for R18 on AffectNet-8. With the application of FAML and T2V, the accuracies of these difficult expressions have been improved respectively on their datasets. Moreover, FAML and T2V have improved accuracy in almost all categories, which demonstrates our models are comprehensive in improving accuracy.

In addition, the number of networks involved in the ensemble is also a crucial factor affecting the classification performance. Therefore, we examine the number of networks from 2 to 6 on RAF-DB and AffectNet-8, whose results of R18+FAML-FGA-T2V are shown in Table \ref{table11}.
\begin{table}[ht!]

\begin{center}
\caption{The evaluation (\%) of different number of models for R18+FAML-FGA-T2V on RAF-DB and AffectNet-8. }
\label{table11}
\setlength{\tabcolsep}{5pt}
%\begin{tabular}{p{55pt}p{55pt}p{190pt}p{100pt}}
\begin{tabular}{cccccc}
\hline
Numbers of Nets         & 2     & 3     & 4     & 5     & 6       \\
\hline
RAF-DB          & 90.58 & 90.84 & 91.30 & 91.46 & \textbf{91.59}    \\
AffectNet-8     & 62.49 & 62.47 & 62.64 & 62.99 & \textbf{63.27}    \\
\hline
\end{tabular}
\end{center}
\end{table}

\begin{figure}[ht!]
            \centering
            \includegraphics[width=0.49\columnwidth]{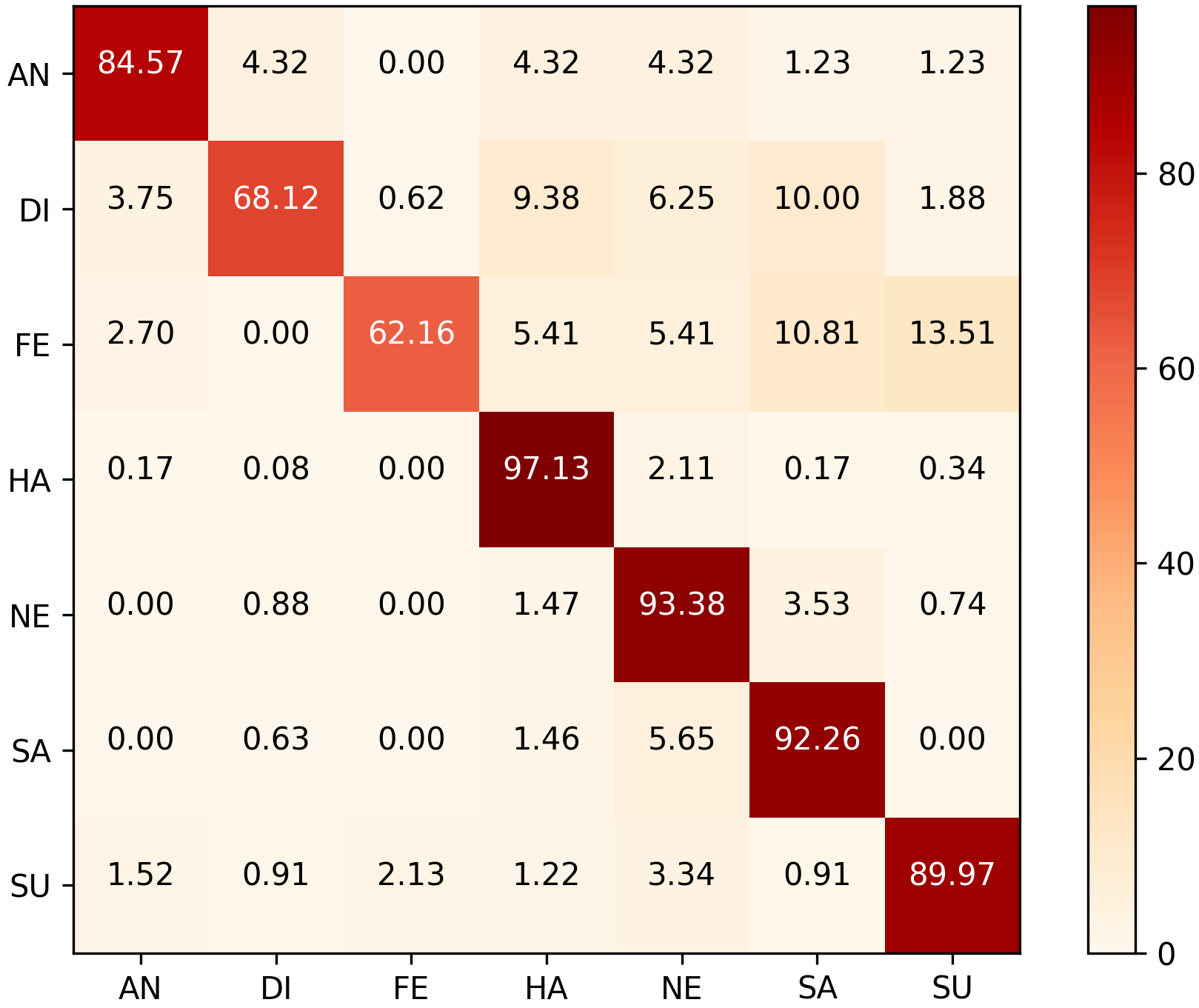}
            \caption{The confusion matrix (\%) of R18+FAML-FGA-T2V with 8 networks participating in ensemble on RAF-DB.}
            \label{fig151}
\end{figure}

From Table \ref{table11}, with the increase in the number of networks, the accuracy of R18+FAML-FGA-T2V shows an upward trend. It can be predicted that when the number of networks increases again, the accuracy will continue to improve. When utilizing 6 networks, R18+FAML-FGA-T2V achieves the state-of-the-art performance $\left(91.59\%, 63.27\%\right)$ on RAF-DB and AffectNet-8. In fact, when increasing the number of networks participating in the ensemble to 8, we can get an ensemble model of R18+FAML-FGA-T2V (8) with \textbf{91.75\%} accuracy on RAF-DB, whose confusion matrix is shown as Fig. \ref{fig151}.
The continuous improvement of accuracy shows that our strategy has more potential.

\subsection{The Comparison with the State-of-the-Art}

\begin{table*}[ht!]
\begin{small}
\begin{center}
\caption{The performance comparison (\%) with state-of-the-art
methods on RAF-DB, AffectNet-8 and  AffectNet-7.}
\label{table111}
\setlength{\tabcolsep}{3pt}
%\begin{tabular}{p{55pt}p{55pt}p{190pt}p{100pt}}
\begin{tabular}{cccccc}
\hline
Types &Model&  RAF-DB&  AffectNet-8 &    AffectNet-7 & Property     \\
\hline
\multirow{14}{*}{Baselines}
&EfficientFace \cite{f46} & 88.36 & 59.89 & 63.70&\multirow{8}{*}{Single} \\
&MA-NET \cite{f25} & 88.04 & 60.29 &64.53\\

&SCN \cite{f47} & 88.14 &60.23& - \\
&KTN \cite{f35} & 88.07 & - &63.97\\
&DACL \cite{f45} & $87.78$ & - & $65.20$ \\

&ADC-Net \cite{f16} & 88.46 & - & - \\

&DAN \cite{f48} & 89.70 &62.09&65.69\\
%MVT \cite{} & 88.62 &-& 64.57 \\
&TransFER \cite{f17} & $90.91$ & -  & $66.23$\\
\cline{2-6}

&GLFEN-9 \cite{e1} &84.72&-&-&\multirow{6}{*}{Ensemble}\\
&SMResNet-WEF \cite{e2} &88.46&-&-\\
&DCNN-SEWP \cite{e4} &87.13&-&-\\
&FLEPNet \cite{e5} &87.56&-&-\\
&WDEA-9-DS \cite{e6} &-&59.8&-\\
&OAENet \cite{f5} & 86.5 & 58.7 &-\\
\hline
\multirow{4}{*}{Ours}
&FAML & $90.32$ & $\textbf{62.17}$ & $65.83$&\multirow{2}{*}{Single}\\
&FAML-FGA & $90.51$ & $\textbf{62.29}$ & $65.89$\\
\cline{2-6}
%FAML-FGA-NOI & $\textbf{91.26}$ & $\textbf{62.69}$ & $\textbf{66.29}$\\
&FAML-FGA-T2V(6)  & $\textbf{91.59}$ & $\textbf{63.27}$ & $\textbf{66.63}$&\multirow{2}{*}{Ensemble}\\
&FAML-FGA-T2V(8)  & $\textbf{91.75}$ & - & -\\
\hline
\end{tabular}
\end{center}
\end{small}
\end{table*}

Table \ref{table111} compares our proposed models to the state-of-the-art methods on RAF-DB, AffectNet-8 and AffectNet-7. For the sake of comprehensiveness, Table \ref{table111} not only compares the performance of single networks, but also compares other ensemble models used for FER.

RAF-DB is a difficult FER dataset and the accuracy of most existing models is between $\left(80\%, 89\%\right)$.
Our proposed  R18+FAML-FGA-T2V is one of the first models to achieve accuracy over 91\% on RAF-DB, which is 0.68\% better than TransFer (90.91\%) \cite{f17}, the best result reported before. Our single models that R18+FAML without ensemble and R18+FAML-FGA with feature fusion among multiple networks respectively achieve 90.32\% and 90.51\% are the only non-ensemble models over 90\% except TransFer \cite{f17}.

AffectNet is the largest and most challenging FER dataset in the wild. The accuracy of most existing models on AffectNet-8 is lower than 62\% except DAN achieved 62.09\% \cite{f48}. Our proposed model R18+FAML-FGA-T2V outperforms the previous best result DAN \cite{f48} by 1.18\%, which is one of the first models to achieve accuracy over 63\% on AffectNet-8. For AffectNet-7, the previous best result is $66.23\%$ achieved by TransFer \cite{f17} followed by DAN \cite{f48} and DACL \cite{f45}. Our proposed model R18+FAML-FGA-T2V achieves $66.63\%$ higher than the state-of-the-art. The single model R18+FAML achieves $65.83\%$ still better than DACL \cite{f45} and DAN \cite{f48} although lower than TransFer \cite{f17}.

Other existing ensemble state-of-the-art models for FER include GLFEN-9 \cite{e1}, SMResNet-WEF \cite{e2}, DCNN-SEWP \cite{e4}, FLEPNet \cite{e5}, WDEA-9-DS \cite{e6}, OAENet \cite{f5}, etc. The best existing ensemble model among them on RAFDB dataset is SMResNet-WEF \cite{e2} with $88.46\%$ accrucy. Our model with $91.59\%$ enhances the accuracy by $3.13\%$. The best existing ensemble model on Affect-8 dataset is WDEA-9-DS \cite{e6} with $59.8\%$ accuracy. Our model with $63.27$ accuracy enhances that by $3.47\%$. This demonstrates that our model still has the highest accuracy even compared with existing ensemble models.

In this paper, we used distributed systems with 3-4 isomorphic GPUs to reduce training time. Devoting to our intentions that providing more possibility to enhance the accuracy of FER and other DL tasks, we didn't consider complexity in this paper. 6-8 networks, resting with our distributed systems (3-4 isomorphic GPUs), enough to exert intrinsic statistical characteristics of fusion,  benefit an efficient research period.
In practical application, we could choose whether to use ensemble strategy and the number of basic networks according to the comprehensive benefits affected by model accuracy and computational complexity. From Table \ref{table11}, the ensemble with only two networks can still effectively improve the performance. Compared with other ensembles between models with different structures \cite{f39,f40,f20}, the advantage of R18+FAML-FGA-T2V is that they are based on the ensemble of networks with the same structure hence approximate training speed, which enables each basic network to be trained in parallel to reduce the time in data reading and preprocessing.

According to the comparison with the existing methods, we not only propose the model R18+FAML-FGA-T2V with ensemble strategies achieving state-of-the-art performance but also the single network R18+FAML without ensemble strategy achieving appreciable performance, which indicates that in addition to our ensemble strategy, the structure and strategies of the single model (combining ResNet18 and FAML) still have certain competitiveness.

\subsection{Universality Studies of T2V Compared with Other Ensembles}

Because improvement of models using T2V involves some general properties of the deep neural network, we believe that T2V has certain universality for other network structures. In order to verify this, we carry out comparative experiments on RAF-DB using some popular network architectures including RI18, ResNet34, VGG16, DenseNet121 and R18 as the basic networks. Adding the R18+FAML and R18+FAML-FGA as basic networks, the results are shown as Table \ref{table12}.  Table \ref{table12} also adds the ensemble performances of T1V (top one voting, i.e., bagging) and NOI (normalized output integrating) as a comparison to demonstrate the superiority of T2V. NOI directly adds the normalized outputs of multiple independently trained models with the same architecture and then classifies. Assuming the output vector of the $l$-th network is $Out^l$, then the NOI output with `Softmax' as normalization is 
\begin{equation}\label{eq8}
Out^{NOI} = \sum\limits^{M}_{l=1} {\rm Softmax}\left({Out^l}\right)
\end{equation}

\begin{table}[ht!]
\begin{small}
\begin{center}
\caption{The evaluation (\%) of different network architectures respectively using T2V, NOI and Bagging on RAF-DB dataset. }
\label{table12}
\setlength{\tabcolsep}{10pt}
%\begin{tabular}{p{55pt}p{55pt}p{190pt}p{100pt}}
\begin{tabular}{ccccc}
\hline
Basic Networks& Origin &  T2V & Bagging & NOI        \\
\hline
RI18 & 85.40 &  88.69 & 87.94& 88.53\\
ResNet34 & 85.53   & 89.24 & 88.85& 89.21\\
VGG16 & 83.83  & 87.52 & 86.67&  87.39\\
DenseNet121 & 86.57   & 89.60 & 89.11& 89.41\\
R18 & 88.62  & 90.42 & 89.73& 90.38\\
\hline
R18+FAML & 90.32 &  91.46 & 90.97& 91.17\\
R18+FAML-FGA & 90.51  & \textbf{91.59} & 91.19& 91.26\\
\hline
\end{tabular}
\end{center}
\end{small}
\end{table}

\begin{figure}[ht!]
            \centering
            \includegraphics[width=0.69\columnwidth]{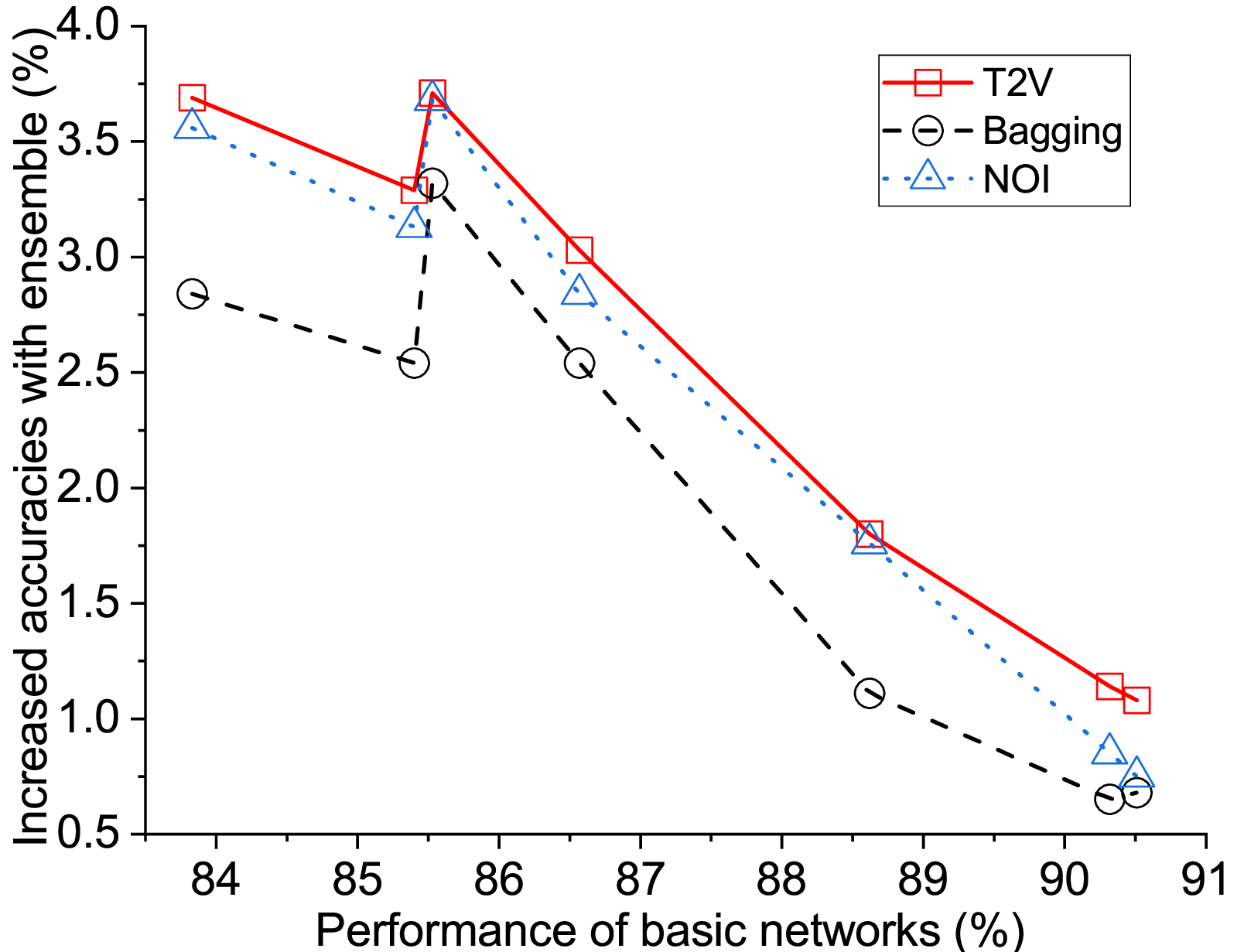}
            \caption{Increased accuracies respectively using ensemble strategy T2V, NOI and Bagging based on different basic networks on RAF-DB dataset. }
            \label{fig13}
\end{figure}

From Table \ref{table12},  T2V improves the performance of RI18, ResNet34, VGG16, DenseNet121 and R18 by $3.29\%$, $3.68\%$, $3.56\%$, $3.03 \%$ and $1.80\%$. When the performance of the basic networks exceeds 90\%, the increases of the performances decrease to $1.14\%$ in R18+FAML and $1.08\%$ in R18+FAML-FGA.
These results demonstrate that our proposed T2V is also effective for networks with other structures hence certain universality.
In order to observe the trend of increase brought by T2V, we take the accuracy of basic networks as the abscissa value and the amplitude of T2V, NOI and T1V as the ordinate value, then we obtain Fig. \ref{fig13}.

From Fig. \ref{fig13}, we can observe that with the increase in the performance of basic networks, the amplitude effect of the ensemble strategies first increases and then decreases. When the accuracy of the basic network is about 85.53\%, the amplitude of T2V is the largest, reaching $ 3.68\%$. Comparing T2V with T1V and NOI, T2V has a better effect on the improvement of accuracies than T1V in the instances of all the examined basic networks, which illustrates that considering multi-level ranking can provide more comprehensive support for FER.

\subsection{Attention Visualization}
To further investigate the effectiveness of our strategies, we leverage the method \cite{f44} to visualize the attention maps generated by the models listed in Table \ref{table10}. We resize the visualization attention maps to the same size as the input images and visualize the attention maps through cv2.COLORMAP\_JET color mapping to the input image. Because the ensemble strategy is leveraged for the final decision in R18+FAML-FGA-T2V with multi networks unable to support gradient backward, we superimpose the attention maps of each network participating in the ensemble according to Equation (\ref{eq9}) to describe the attention maps of ensemble models.

\begin{figure}[ht!]
            \centering
            \includegraphics[width=0.99\columnwidth]{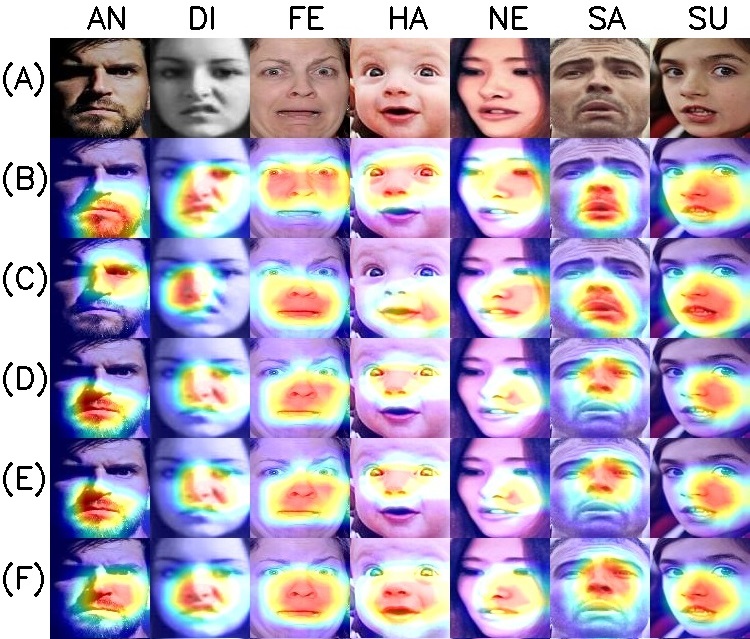}
            \caption{Attention visualization \cite{f44} of different expressions on some example face images from RAF-DB dataset. (A)-(F) respectively denote (A) input image, (B) R18, (C) R18+FA, (D) R18+FAML, (E) R18+FAML-FGA and (F) R18+FAML-FGA-T2V. }
            \label{fig15}
\end{figure}

Fig. \ref{fig15} plots the attention maps of different categories of expressions on RAF-DB dataset. Fig. \ref{fig15} consists $6\times 7$ sub-figures. Each column is one of the seven categories of expression. From left to right, the expression categories are anger (AN), disgust (DI), fear (FE), happiness(HA), neutral (NE), sadness (SA) and surprise (SU). The first row marked as (A) shows the original input image, and the second to sixth rows marked from (B) to (F) shows the result of five models corresponding to that in Table \ref{table10}: (B) R18; (C) R18+FA; (D) R18+FAML; (E) R18+FAML-FGA; (F) R18+FAML-FGA-T2V.

Comparing different rows (training strategies or decision strategies): the areas concerned by R18+FA (C) are different from those of the baseline R18 (B) and their areas both have a lower correlation with expression than other models;
when adding multiple loss functions including LSR and CE loss, R18+FAML (D) slightly reduces the scope of interest areas but corrects the area to the location with a higher correlation with expression; R18+FAML-FGA (E) is similar to R18+FAML (D);
and T2V (F) corrects not only the location but also adjusts (including expanding and reducing) the scope of interest areas.
Finally, T2V obtains the interest area with a higher correlation by removing excess areas and expanding expression-aware areas. For expanding expression-aware area, E.g. (B), (D) and (E) in happiness include eyes and nose but without mouth; (C) of that includes nose and mouth but without eyes; (F) using ensemble strategy covers mouth, nose and eyes.
For removing excess areas, E.g. (F) in surprise remove the excess area of cheeks compared with (B) and remove the excess area of chin compared with (C).

Comparing different columns (expressions): although the mouth, nose and eyes are generally assumed as the most significant areas to recognize expressions, they may be similar for even different expressions \cite{f17}. In this case, our proposed models conduct recognition with the assistance of other areas or by focusing on a certain area. For example: disgusts in (D) to (F)  focus on the upper lip and nasolabial sulcus; surprises in (D) to (F) take the forehead into the interest area.

Fig. \ref{fig15} shows that our proposed strategies including FAML, FGA and T2V demonstrate the ability to concentrate interest on key areas of expression, which active a mechanism analogue to attention mechanism. In fact, we conducted tests and found that without the addition of attention blocks in backbones, our proposed models can still achieve a high level of proficiency, even achieving the same accuracy as that with attention blocks in RAF-DB and AffectNet datasets.

\section{Conclusion and Future work}\label{sec5}
For the FER task, we firstly propose a novel network architecture R18+FAML by using ResNet18 as the backbone, adding three attention blocks respectively after the first three residual blocks, adding an internal feature fusion block, and using multi loss functions including LSR loss and CE loss to train the network. With the above strategies, R18+FAML can learn the expression information of the key areas and has a high-level performance on three challenging unbalanced FER datasets.
Secondly, we propose feature fusion among networks based on the genetic algorithm (FGA) making use of the operational characteristics of convolution and effectively combining the feature extractors (i.e. weights of convolution kernels) of multiple networks, which results in the improvement of FER.
To further improve the accuracy of FER, we finally propose one ensemble decision-making strategy, i.e. T2V (Top Two Voting), which can obviously enhance the accuracy of various networks.
Combining R18+FAML, FGA and the T2V, we obtain an effective ensemble model R18+FAML-FGA-T2V, which can focus on the areas with higher relevance to expressions.

Extensive experiments not only demonstrate the effectiveness of our proposed strategies for FER but also show our models outperform the state-of-the-art models. Specifically, our proposed single model R18+FAML achieves $\left( 90.32, 62.17, 65.83 \right)\%$ and our proposed ensemble model R18+FAML-FGA-T2V achieves the accuracies of $\left( 91.59, 63.27, 66.63 \right)\%$ respectively on RAF-DB, AffectNet-8 and  AffectNet-7. To the best of our knowledge, our proposed models are one of the first models with accuracies over $91\%$ on RAF-DB and over $63\%$ on AffectNet-8.

In future work, we plan to adjust the feature fusion strategies and convolution operation mode to explore more effective feature extraction methods for FER. Additionally, we will study the strategies to reduce the computation complexity such as automatic parameter adjustment based on reinforcement learning or adaptive learning. In addition, our proposed strategies can be applied to other well-performed models (e.g. TransFER \cite{f17}) to create FER models with ultra-highly accuracy.

\section*{Acknowledgments}
This research is partially supported  by the Project of Key Research and Development Program of Sichuan Province with Grant ID 2021YFG0325, {and by the Chengdu Science and Technology Project with Grant ID 2022-YF05-02014-SN}.

\bibliographystyle{elsarticle-num}
\bibliography{mybibfile}

\begin{thebibliography}{10}
\expandafter\ifx\csname url\endcsname\relax
  \def\url#1{\texttt{#1}}\fi
\expandafter\ifx\csname urlprefix\endcsname\relax\def\urlprefix{URL }\fi
\expandafter\ifx\csname href\endcsname\relax
  \def\href#1#2{#2} \def\path#1{#1}\fi

\bibitem{f2}
S.~Li, W.~Li, S.~Wen, K.~Shi, Y.~Yang, P.~Zhou, T.~Huang, Auto-fernet: {A}
  facial expression recognition network with architecture search, {IEEE} Trans.
  Netw. Sci. Eng. 8~(3) (2021) 2213--2222.

\bibitem{f27}
J.~Chen, C.~Guo, R.~Xu, K.~Zhang, Z.~Yang, H.~Liu, Toward children's empathy
  ability analysis: Joint facial expression recognition and intensity
  estimation using label distribution learning, {IEEE} Trans. Ind. Informatics
  18~(1) (2022) 16--25.

\bibitem{f3}
X.~Zhao, J.~Zhu, B.~Luo, Y.~Gao, Survey on facial expression recognition:
  History, applications, and challenges, {IEEE} Multim. 28~(4) (2021) 38--44.

\bibitem{f4}
M.~Li, H.~Xu, X.~Huang, Z.~Song, X.~Liu, X.~Li, Facial expression recognition
  with identity and emotion joint learning, {IEEE} Trans. Affect. Comput.
  12~(2) (2021) 544--550.

\bibitem{f5}
Z.~Wang, F.~Zeng, S.~Liu, B.~Zeng, Oaenet: Oriented attention ensemble for
  accurate facial expression recognition, Pattern Recognit. 112 (2021) 107694.

\bibitem{f6}
Q.~Huang, C.~Huang, X.~Wang, F.~Jiang, Facial expression recognition with
  grid-wise attention and visual transformer, Inf. Sci. 580 (2021) 35--54.

\bibitem{f7}
Y.~Li, J.~Yang, Y.~Song, L.~Cao, J.~Luo, L.~Li, Learning from noisy labels with
  distillation, in: {IEEE} International Conference on Computer Vision, {ICCV}
  2017, Venice, Italy, October 22-29, 2017, {IEEE} Computer Society, 2017, pp.
  1928--1936.

\bibitem{f10}
Z.~Zhang, P.~Luo, C.~C. Loy, X.~Tang, From facial expression recognition to
  interpersonal relation prediction, Int. J. Comput. Vis. 126~(5) (2018)
  550--569.

\bibitem{f9}
C.~F. Benitez{-}Quiroz, R.~Srinivasan, A.~M. Mart{\'{\i}}nez, Emotionet: An
  accurate, real-time algorithm for the automatic annotation of a million
  facial expressions in the wild, in: 2016 {IEEE} Conference on Computer Vision
  and Pattern Recognition, {CVPR} 2016, Las Vegas, NV, USA, June 27-30, 2016,
  {IEEE} Computer Society, 2016, pp. 5562--5570.

\bibitem{f8}
A.~Mollahosseini, B.~Hassani, M.~H. Mahoor, Affectnet: {A} database for facial
  expression, valence, and arousal computing in the wild, {IEEE} Trans. Affect.
  Comput. 10~(1) (2019) 18--31.

\bibitem{f1}
L.~Liang, C.~Lang, Y.~Li, S.~Feng, J.~Zhao, Fine-grained facial expression
  recognition in the wild, {IEEE} Trans. Inf. Forensics Secur. 16 (2021)
  482--494.

\bibitem{f11}
K.~Simonyan, A.~Zisserman, Very deep convolutional networks for large-scale
  image recognition, in: Y.~Bengio, Y.~LeCun (Eds.), 3rd International
  Conference on Learning Representations, {ICLR} 2015, San Diego, CA, USA, May
  7-9, 2015, Conference Track Proceedings, 2015.

\bibitem{f12}
K.~He, X.~Zhang, S.~Ren, J.~Sun, Deep residual learning for image recognition,
  in: 2016 {IEEE} Conference on Computer Vision and Pattern Recognition, {CVPR}
  2016, Las Vegas, NV, USA, June 27-30, 2016, {IEEE} Computer Society, 2016,
  pp. 770--778.

\bibitem{f13}
G.~Huang, Z.~Liu, L.~van~der Maaten, K.~Q. Weinberger, Densely connected
  convolutional networks, in: 2017 {IEEE} Conference on Computer Vision and
  Pattern Recognition, {CVPR} 2017, Honolulu, HI, USA, July 21-26, 2017, {IEEE}
  Computer Society, 2017, pp. 2261--2269.

\bibitem{f14}
Q.~Zhu, L.~Gao, H.~Song, Q.~Mao, Learning to disentangle emotion factors for
  facial expression recognition in the wild, Int. J. Intell. Syst. 36~(6)
  (2021) 2511--2527.

\bibitem{f15}
T.~Rao, J.~Li, X.~Wang, Y.~Sun, H.~Chen, Facial expression recognition with
  multiscale graph convolutional networks, {IEEE} Multim. 28~(2) (2021) 11--19.

\bibitem{f16}
H.~Xia, C.~Li, Y.~Tan, L.~Li, S.~Song, Destruction and reconstruction learning
  for facial expression recognition, {IEEE} Multim. 28~(2) (2021) 20--28.

\bibitem{f17}
F.~Xue, Q.~Wang, G.~Guo, Transfer: Learning relation-aware facial expression
  representations with transformers, in: Proceedings of the IEEE/CVF
  International Conference on Computer Vision, {IEEE} Computer Society, 2021,
  pp. 3601--3610.

\bibitem{f48}
Z.~Wen, W.~Lin, T.~Wang, G.~Xu, Distract your attention: Multi-head cross
  attention network for facial expression recognition, CoRR abs/2109.07270
  (2021).

\bibitem{f25}
Z.~Zhao, Q.~Liu, S.~Wang, Learning deep global multi-scale and local attention
  features for facial expression recognition in the wild, {IEEE} Trans. Image
  Process. 30 (2021) 6544--6556.

\bibitem{f24}
Q.~Wang, J.~Lai, L.~Claesen, Z.~Yang, L.~Lei, W.~Liu, A novel feature
  representation: Aggregating convolution kernels for image retrieval, Neural
  Networks 130 (2020) 1--10.

\bibitem{f23}
L.~Yang, Z.~Wang, S.~Gao, M.~Shi, B.~Liu, Magnetic flux leakage image
  classification method for pipeline weld based on optimized convolution
  kernel, Neurocomputing 365 (2019) 229--238.

\bibitem{f18}
G.~Wen, Z.~Hou, H.~Li, D.~Li, L.~Jiang, E.~Xun, Ensemble of deep neural
  networks with probability-based fusion for facial expression recognition,
  Cogn. Comput. 9~(5) (2017) 597--610.

\bibitem{f19}
H.~He, S.~Chen, Identification of facial expression using a multiple impression
  feedback recognition model, Appl. Soft Comput. 113~(Part) (2021) 107930.

\bibitem{f20}
N.~K. Benamara, M.~Val{-}Calvo, J.~R.~{\'{A}}. S{\'{a}}nchez,
  A.~D{\'{\i}}az{-}Morcillo, J.~M.~F. de~Vicente, E.~Fern{\'{a}}ndez{-}Jover,
  T.~B. Stambouli, Real-time facial expression recognition using smoothed deep
  neural network ensemble, Integr. Comput. Aided Eng. 28~(1) (2021) 97--111.

\bibitem{f21}
V.~R.~R. Chirra, S.~R. Uyyala, V.~K.~K. Kolli, Virtual facial expression
  recognition using deep {CNN} with ensemble learning, J. Ambient Intell.
  Humaniz. Comput. 12~(12) (2021) 10581--10599.

\bibitem{f22}
H.~Sikkandar, R.~Thiyagarajan, Deep learning based facial expression
  recognition using improved cat swarm optimization, J. Ambient Intell.
  Humaniz. Comput. 12~(2) (2021) 3037--3053.

\bibitem{f28}
D.~Poux, B.~Allaert, N.~Ihaddadene, I.~M. Bilasco, C.~Djeraba, M.~Bennamoun,
  Dynamic facial expression recognition under partial occlusion with optical
  flow reconstruction, {IEEE} Trans. Image Process. 31 (2022) 446--457.

\bibitem{f29}
X.~Jin, Z.~Jin, Miniexpnet: {A} small and effective facial expression
  recognition network based on facial local regions, Neurocomputing 462 (2021)
  353--364.

\bibitem{f30}
S.~Xie, H.~Hu, Y.~Chen, Facial expression recognition with two-branch
  disentangled generative adversarial network, {IEEE} Trans. Circuits Syst.
  Video Technol. 31~(6) (2021) 2359--2371.

\bibitem{f31}
W.~Xie, L.~Shen, J.~Duan, Adaptive weighting of handcrafted feature losses for
  facial expression recognition, {IEEE} Trans. Cybern. 51~(5) (2021)
  2787--2800.

\bibitem{f49}
G.~Karimi, M.~Heidarian, Facial expression recognition with polynomial legendre
  and partial connection {MLP}, Neurocomputing 434 (2021) 33--44.

\bibitem{f50}
T.~Ma, W.~Tian, Y.~Xie, Multi-level knowledge distillation for low-resolution
  object detection and facial expression recognition, Knowl. Based Syst. 240
  (2022) 108136.

\bibitem{f33}
X.~Sun, P.~Xia, F.~Ren, Multi-attention based deep neural network with hybrid
  features for dynamic sequential facial expression recognition, Neurocomputing
  444 (2021) 378--389.

\bibitem{k2}
A.~R. Shahid, H.~Yan, Squeezexpnet: Dual-stage convolutional neural network for
  accurate facial expression recognition with attention mechanism, Knowl. Based
  Syst. 269 (2023) 110451.

\bibitem{k4}
H.~Li, H.~Xu, Deep reinforcement learning for robust emotional classification
  in facial expression recognition, Knowl. Based Syst. 204 (2020) 106172.

\bibitem{k3}
F.~Nan, W.~Jing, F.~Tian, J.~Zhang, K.~Chao, Z.~Hong, Q.~Zheng, Feature
  super-resolution based facial expression recognition for multi-scale
  low-resolution images, Knowl. Based Syst. 236 (2022) 107678.

\bibitem{f32}
Z.~Li, C.~Wang, X.~Liu, Y.~Wang, Facial expression description and recognition
  based on fuzzy semantic concepts, Future Gener. Comput. Syst. 114 (2021)
  619--628.

\bibitem{f34}
B.~Chen, W.~Guan, P.~Li, N.~Ikeda, K.~Hirasawa, H.~Lu, Residual multi-task
  learning for facial landmark localization and expression recognition, Pattern
  Recognit. 115 (2021) 107893.

\bibitem{f35}
H.~Li, N.~Wang, X.~Ding, X.~Yang, X.~Gao, Adaptively learning facial expression
  representation via {C-F} labels and distillation, {IEEE} Trans. Image
  Process. 30 (2021) 2016--2028.

\bibitem{f36}
H.~Ghazouani, A genetic programming-based feature selection and fusion for
  facial expression recognition, Appl. Soft Comput. 103 (2021) 107173.

\bibitem{f37}
S.~Lin, M.~Bai, F.~Liu, L.~Shen, Y.~Zhou, Orthogonalization-guided feature
  fusion network for multimodal 2d+3d facial expression recognition, {IEEE}
  Trans. Multim. 23 (2021) 1581--1591.

\bibitem{f38}
M.~Sui, Z.~Zhu, F.~Zhao, F.~Wu, Ffnet-m: Feature fusion network with masks for
  multimodal facial expression recognition, in: 2021 {IEEE} International
  Conference on Multimedia and Expo, {ICME} 2021, Shenzhen, China, July 5-9,
  2021, {IEEE}, 2021, pp. 1--6.

\bibitem{f39}
A.~Renda, M.~Barsacchi, A.~Bechini, F.~Marcelloni, Comparing ensemble
  strategies for deep learning: An application to facial expression
  recognition, Expert Syst. Appl. 136 (2019) 1--11.

\bibitem{f40}
W.~Sun, H.~Zhao, Z.~Jin, A facial expression recognition method based on
  ensemble of 3d convolutional neural networks, Neural Comput. Appl. 31~(7)
  (2019) 2795--2812.

\bibitem{e3}
R.~Wadhawan, T.~K. Gandhi, Landmark-aware and part-based ensemble transfer
  learning network for static facial expression recognition from images, {IEEE}
  Trans. Artif. Intell. 4~(2) (2023) 349--361.

\bibitem{e4}
J.~Y. Choi, B.~Lee, Combining deep convolutional neural networks with
  stochastic ensemble weight optimization for facial expression recognition in
  the wild, {IEEE} Trans. Multim. 25 (2023) 100--111.

\bibitem{e5}
K.~Mohan, A.~Seal, A.~Yazidi, O.~Krejcar, Flepnet: Feature level ensemble
  parallel network for facial expression recognition, {IEEE} Trans. Affect.
  Comput. 13~(4) (2022) 2058--2070.

\bibitem{f41}
A.~Dosovitskiy, L.~Beyer, A.~Kolesnikov, D.~Weissenborn, X.~Zhai,
  T.~Unterthiner, M.~Dehghani, M.~Minderer, G.~Heigold, S.~Gelly, J.~Uszkoreit,
  N.~Houlsby, An image is worth 16x16 words: Transformers for image recognition
  at scale, in: 9th International Conference on Learning Representations,
  {ICLR} 2021, Virtual Event, Austria, May 3-7, 2021, OpenReview.net, 2021.

\bibitem{f43}
Y.~Guo, L.~Zhang, Y.~Hu, X.~He, J.~Gao, Ms-celeb-1m: {A} dataset and benchmark
  for large-scale face recognition, in: B.~Leibe, J.~Matas, N.~Sebe, M.~Welling
  (Eds.), Computer Vision - {ECCV} 2016 - 14th European Conference, Amsterdam,
  The Netherlands, October 11-14, 2016, Proceedings, Part {III}, Vol. 9907 of
  Lecture Notes in Computer Science, Springer, 2016, pp. 87--102.

\bibitem{f46}
Z.~Zhao, Q.~Liu, F.~Zhou, Robust lightweight facial expression recognition
  network with label distribution training, in: Thirty-Fifth {AAAI} Conference
  on Artificial Intelligence, {AAAI} 2021, Thirty-Third Conference on
  Innovative Applications of Artificial Intelligence, {IAAI} 2021, The Eleventh
  Symposium on Educational Advances in Artificial Intelligence, {EAAI} 2021,
  Virtual Event, February 2-9, 2021, {AAAI} Press, 2021, pp. 3510--3519.

\bibitem{f47}
K.~Wang, X.~Peng, J.~Yang, S.~Lu, Y.~Qiao, Suppressing uncertainties for
  large-scale facial expression recognition, in: 2020 {IEEE/CVF} Conference on
  Computer Vision and Pattern Recognition, {CVPR} 2020, Seattle, WA, USA, June
  13-19, 2020, Computer Vision Foundation / {IEEE}, 2020, pp. 6896--6905.

\bibitem{f45}
A.~H. Farzaneh, X.~Qi, Facial expression recognition in the wild via deep
  attentive center loss, in: {IEEE} Winter Conference on Applications of
  Computer Vision, {WACV} 2021, Waikoloa, HI, USA, January 3-8, 2021, {IEEE},
  2021, pp. 2401--2410.

\bibitem{e1}
Z.~He, B.~Meng, L.~Wang, G.~Jeon, Z.~Liu, X.~Yang, Global and local fusion
  ensemble network for facial expression recognition, Multim. Tools Appl.
  82~(4) (2023) 5473--5494.

\bibitem{e2}
J.~Liu, M.~Hu, Y.~Wang, Z.~Huang, J.~Jiang, Symmetric multi-scale residual
  network ensemble with weighted evidence fusion strategy for facial expression
  recognition, Symmetry 15~(6) (2023) 1228.

\bibitem{e6}
G.~Han, C.~Chen, Z.~Xu, S.~Zhou, Weighted ensemble with angular feature
  learning for facial expression recognition, J. Intell. Fuzzy Syst. 41~(6)
  (2021) 6845--6857.

\bibitem{f44}
H.~Chefer, S.~Gur, L.~Wolf, Transformer interpretability beyond attention
  visualization, in: {IEEE} Conference on Computer Vision and Pattern
  Recognition, {CVPR} 2021, virtual, June 19-25, 2021, Computer Vision
  Foundation / {IEEE}, 2021, pp. 782--791.

\end{thebibliography}

\end{sloppypar}

\end{document}